\documentclass[11pt]{article}

\usepackage[preprint]{acl}

\usepackage{times}
\usepackage{latexsym}

\usepackage[T1]{fontenc}

\usepackage[utf8]{inputenc}

\usepackage{microtype}

\usepackage{inconsolata}

\usepackage{graphicx}

\usepackage{hyperref}
\usepackage{url}

\usepackage[utf8]{inputenc} 
\usepackage[T1]{fontenc}    
\usepackage{hyperref}       
\usepackage{url}            
\usepackage{booktabs}       
\usepackage{amsfonts}       
\usepackage{nicefrac}       
\usepackage{microtype}      
\usepackage{xcolor}         
\usepackage{amsmath}
\usepackage{amssymb}
\usepackage{mathtools}
\usepackage{amsthm}
\usepackage{caption}
\usepackage{graphicx}
\usepackage{multirow}
\usepackage{enumitem}
\usepackage{color}
\usepackage{xcolor}
\usepackage{colortbl}
\usepackage{pifont}
\usepackage{algorithm}
\usepackage{algorithmic}
\usepackage{ulem}
\usepackage{booktabs}
\usepackage{bbding}
\usepackage{wrapfig}
\usepackage[most]{tcolorbox} 
\usepackage[table]{xcolor}
\usepackage{subcaption}
\definecolor{DarkGreen}{RGB}{1,100,32} 
\usepackage[capitalize,noabbrev]{cleveref}

\theoremstyle{plain}

\theoremstyle{definition}

\theoremstyle{remark}

\definecolor{myDarkGreen}{RGB}{68, 126, 102}
\definecolor{myLightGreen}{RGB}{245, 250, 248}

\newtcolorbox{problembox}[1]{
    colback=myLightGreen,      
    colframe=myDarkGreen,      
    colbacktitle=myDarkGreen,  
    coltitle=white,            
    fonttitle=\bfseries,       
    breakable,
    arc=3pt,                   
    left=10pt,                 
    right=10pt,                
    top=10pt,                  
    bottom=10pt,               
    enhanced,                  
    attach boxed title to top left={yshift=-2mm, xshift=2mm}, 
    title={#1},                
    toptitle=3pt,              
    bottomtitle=3pt            
}

\title{AlgBench: To What Extent Do Large Reasoning Models Understand Algorithms?}

\author{
 \textbf{Henan Sun\textsuperscript{1}},
 \textbf{Kaichi Yu\textsuperscript{3}},
 \textbf{Yuyao Wang\textsuperscript{1}},
 \textbf{Bowen Liu\textsuperscript{1}},
\\
 \textbf{Xunkai Li\textsuperscript{3}},
 \textbf{Rong-Hua Li\textsuperscript{3}},
 \textbf{Nuo Chen\textsuperscript{1}},
 \textbf{Jia Li \textsuperscript{1,2,*}},
\\
\\
 \textsuperscript{1}The Hong Kong University of Science and Technology (Guangzhou),\\
 \textsuperscript{2}The Hong Kong University of Science and Technology,\\
 \textsuperscript{3}Beijing Institute of Technology,
\\
 \small{
   \href{mailto:magneto0617@foxmail.com}{magneto0617@foxmail.com},
   \href{mailto:ykckevin01@gmail.com}{ykckevin01@gmail.com},
   \href{mailto:ywang737@connect.hkust-gz.edu.cn}{ywang737@connect.hkust-gz.edu.cn},
   \href{mailto:bliu690@connect.hkust-gz.edu.cn}{bliu690@connect.hkust-gz.edu.cn}
 }
 \\
 \small{
    \href{mailto:cs.xunkai.li@gmail.com}{cs.xunkai.li@gmail.com},
    \href{mailto:lironghuabit@126.com}{lironghuabit@126.com},
    \href{mailto:chennuo26@gmail.com}{chennuo26@gmail.com},
    \href{mailto:jialee@hkust-gz.edu.cn}{jialee@hkust-gz.edu.cn}
 }
}

\begin{document}
\maketitle

\renewcommand{\thefootnote}{} 
\footnotetext{*: Corresponding Author}

\begin{abstract}
Reasoning ability has become a central focus in the advancement of Large Reasoning Models (LRMs). Although notable progress has been achieved on several reasoning benchmarks such as MATH500 and LiveCodeBench, existing benchmarks for algorithmic reasoning remain limited, failing to answer a critical question: \textit{Do LRMs truly master algorithmic reasoning?} To answer this question, we propose AlgBench, an expert-curated benchmark that evaluates LRMs under an \textbf{algorithm-centric paradigm}.

AlgBench consists of over 3,000 original problems spanning 27 algorithms, constructed by ACM algorithmic experts and organized under a comprehensive taxonomy, including Euclidean-structured, non-Euclidean-structured, non-optimized, local-optimized, global-optimized, and heuristic-optimized categories. Empirical evaluations on leading LRMs (e.g., Gemini-3-Pro, DeepSeek-v3.2-Speciale and GPT-o3) reveal substantial performance heterogeneity: while models perform well on non-optimized tasks (up to 92\%), accuracy drops sharply to around 49\% on globally optimized algorithms such as dynamic programming. Further analysis uncovers \textbf{strategic over-shifts}, wherein models prematurely abandon correct algorithmic designs due to \textbf{necessary low-entropy tokens}. These findings expose fundamental limitations of problem-centric reinforcement learning and highlight the necessity of an algorithm-centric training paradigm for robust algorithmic reasoning.
\end{abstract}

\section{Introduction}
\label{sec: intro}
Reasoning ability has emerged as a central concern in both the research and practical deployment of large language models (LLMs)~\cite{chen2025towards}. The development of advanced reasoning-oriented models, such as Gemini-3-pro~\cite{google2025gemini3}, Deepseek-v3.2-speciale~\cite{liu2025deepseek} and OpenAI’s o3~\cite{openai2025reasoning}, underscores the significant potential of this line of inquiry. To facilitate the design of high-performance reasoning LLMs, the construction of high-quality reasoning benchmarks has become increasingly indispensable for both model training and evaluation~\cite{glazer2024frontiermath, gulati2024putnam, he2024olympiadbench, huang2025math, huang2024olympicarena}.

\begin{figure*}[t]
	\centering
  \includegraphics[width=0.8\textwidth]{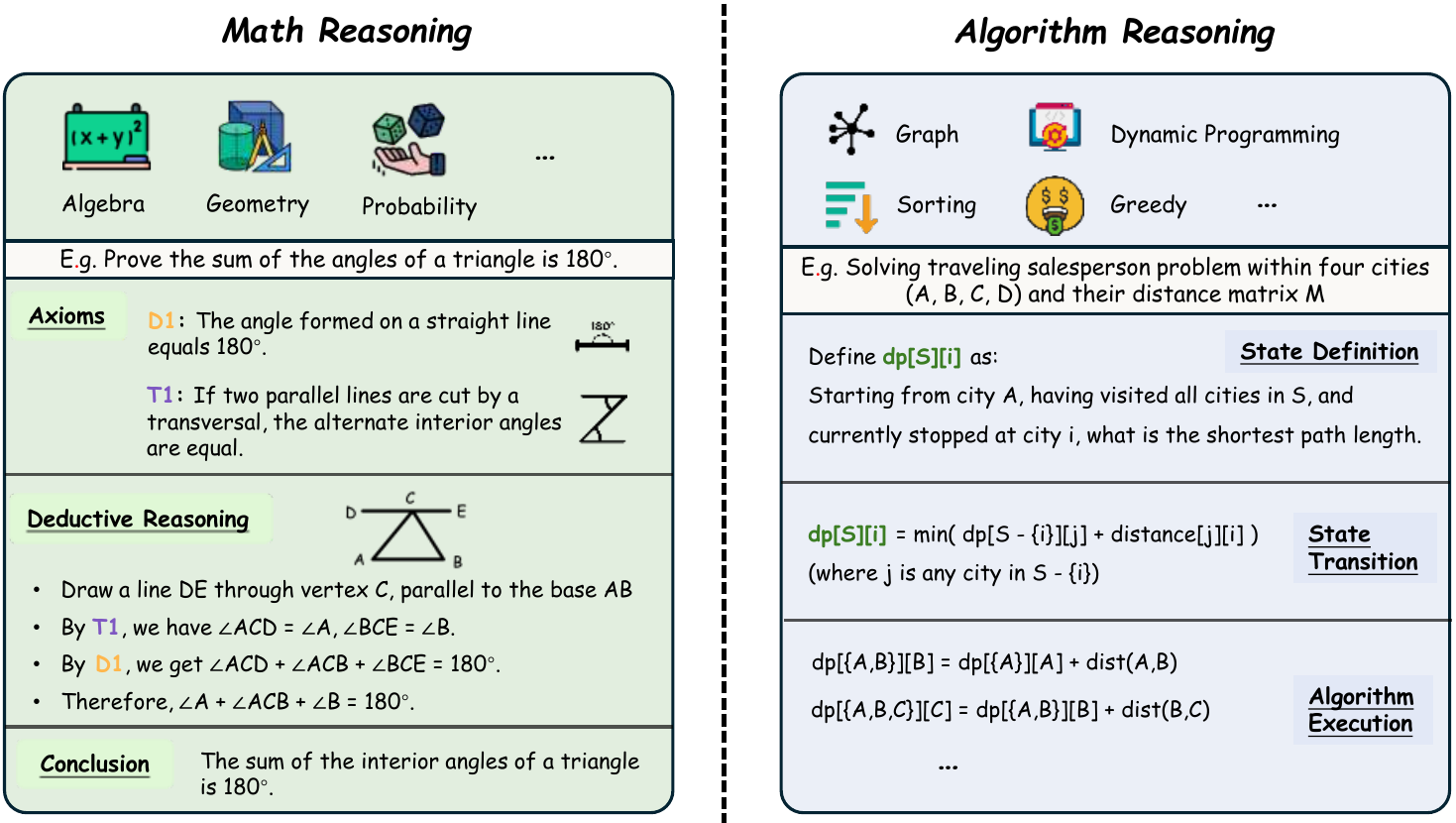}
  \caption{
    The difference between math reasoning and algorithmic reasoning.
}
  \label{fig:math_vs_alg}
\end{figure*}

Significant progress has been made in the construction of reasoning benchmarks, particularly within the field of mathematical reasoning, exemplified by resources such as MATH-500~\cite{hendrycks2021measuring}, AIME 2024~\cite{he2024olympiadbench}, and AIME 2025~\cite{petrov2025proof}. However, the availability of high-quality benchmarks for algorithmic reasoning remains limited, thereby hindering advancements in enhancing the algorithmic reasoning capabilities of LLMs~\cite{li2024glbench, zheng2025livecodebench, yang2025probench, wei2025equibench}. As illustrated in Figure~\ref{fig:math_vs_alg}, algorithmic reasoning skills cannot be acquired merely through training on mathematical reasoning benchmarks, owing to several fundamental distinctions as follows: (1) reasoning paradigms: mathematical reasoning primarily follows a deductive paradigm, whereas algorithmic reasoning is characterized by procedural reasoning. As depicted in Figure~\ref{fig:math_vs_alg}, a classical example of the deductive paradigm is Aristotle’s three-stage reasoning~\cite{mcleod1995aristotle}, which emphasizes axiom construction, theorem deduction, and precise calculation. By contrast, a canonical example of algorithmic reasoning is dynamic programming, which centers on data structuring and the systematic design of algorithmic procedures. (2) reasoning domains: mathematical reasoning is largely concerned with areas such as algebra, geometry, probability and so on, while algorithmic reasoning places greater emphasis on tasks involving sequence sorting, graph theory, dynamic programming and so on~\cite{xia2025leetcodedataset, jain2024livecodebench, zheng2025livecodebench}.
Given these multi-level distinctions between mathematical and algorithmic reasoning, together with the current scarcity of high-quality algorithmic reasoning benchmarks, the development of such benchmarks constitutes an urgent and critical research priority.

Although several algorithmic benchmarks have been introduced, including CLRS-Text~\cite{markeeva2024clrs}, LeetCodeDataset~\cite{xia2025leetcodedataset}, LiveCodeBench~\cite{jain2024livecodebench} and LiveCodeBenchPro~\cite{zheng2025livecodebench}, they still fail to answer a critical question: \textit{whether LRMs truly learn algorithmic reasoning?} More specifically, these benchmarks continue to exhibit several critical limitations shown below:

\textit{Limitation 1}: Existing algorithmic benchmarks frequently draw their problem sets from well-known coding platforms such as CodeForces and LeetCode. However, tasks sourced from these platforms often encompass multiple algorithms simultaneously, giving rise to several challenges. First, it becomes ill-suited for evaluating LRMs' proficiency in a single algorithm, since failure on a multi-algorithm task does not allow researchers to isolate which specific algorithmic reasoning ability is lacking. Second, it is difficult for LLMs to acquire a well-defined algorithmic reasoning paradigm, as individual problems may require the integration of reasoning skills across several algorithms—demands that often exceed the current capabilities of LLMs.

\textit{Limitation 2}: The aforementioned benchmarks encompass only a restricted subset of algorithms. For example, CLRS-Text is limited to several fundamental algorithms presented in the Introduction to Algorithms textbook~\cite{cormen2022introduction}, while omitting a wide range of important algorithms that are commonly utilized in algorithmic competitions and industrial applications, such as tree-based dynamic programming, network flow and so on.

\textit{Limitation 3}: Existing algorithmic benchmarks are highly susceptible to data contamination, as their problem sets are typically derived from publicly available sources such as textbooks and competitive programming websites. This makes it difficult to ensure that LLMs have not been exposed to these problems during pretraining~\cite{white2024livebench, jain2024livecodebench, zheng2025livecodebench}. A parallel case has been observed in the domain of mathematical reasoning, where Wu et al. demonstrated that the Qwen2.5 series likely leveraged data from benchmarks such as MATH and AIME, thereby achieving inflated performance on mathematical reasoning tasks~\cite{shao2025spurious}.

To address the limitations illustrated above, we introduce AlgBench, a novel benchmark for evaluating LRMs' algorithmic reasoning. Figure~\ref{fig:benchmark_overview} presents an overview of our benchmark. To summarize, our contributions are as follow:

\begin{itemize}
    \item \textbf{Paradigm Shift:} Unlike prior benchmarks that adhere to a problem-centric paradigm, we are the first to adopt the \textbf{algorithm-centric paradigm} to the best of our knowledge, which more closely aligns with the mechanisms of human reasoning. As discussed in \textit{Limitation 1}, existing benchmarks grounded in problem-centric paradigms are insufficient to assess whether LRMs genuinely acquire a coherent and transferable understanding of a given algorithm. To address this limitation, we adopt an algorithm-centric paradigm in the construction of our dataset, enabling a more rigorous and systematic evaluation of the algorithmic reasoning capabilities of state-of-the-art LRMs.

    \item \textbf{Novel-taxonomic and Contamination-free Dataset Construction:} We engage multiple algorithmic experts to manually construct problems guided by a novel algorithmic taxonomy, including Euclidean-structured, non-Euclidean-structured, non-optimized, locally optimized, globally optimized, and heuristically optimized categories. The resulting dataset comprises over 3,000 original problems spanning 27 distinct algorithms, thereby substantially enhancing algorithmic diversity while mitigating the risk of data contamination.

    \item \textbf{Novel Research Questions:} We systematically investigate: (1) To what extent do LRMs demonstrate genuine mastery of algorithmic reasoning? (2) What underlying factors contribute to the performance deficits of LRMs in algorithms requiring global optimization strategies? Through a comprehensive evaluation, we identify three salient findings: (1) pronounced heterogeneity in LRM performance across different algorithmic taxonomies; (2) uneven performance gains across these taxonomies under parameter scaling; and (3) the emergence of strategic over-shifts induced by the presence of necessary low-entropy tokens (shown in the lower-right of Figure~\ref{fig:benchmark_overview}).
\end{itemize}

\section{Dataset Construction}
\label{sec: dataset}
Figure~\ref{fig:benchmark_overview} illustrates the construction process of our dataset.
\subsection{Algorithm-centric Paradigm}
\label{sec: alg_centric_paradigm}
\begin{figure*}[t]
	\centering
  \includegraphics[width=\textwidth]{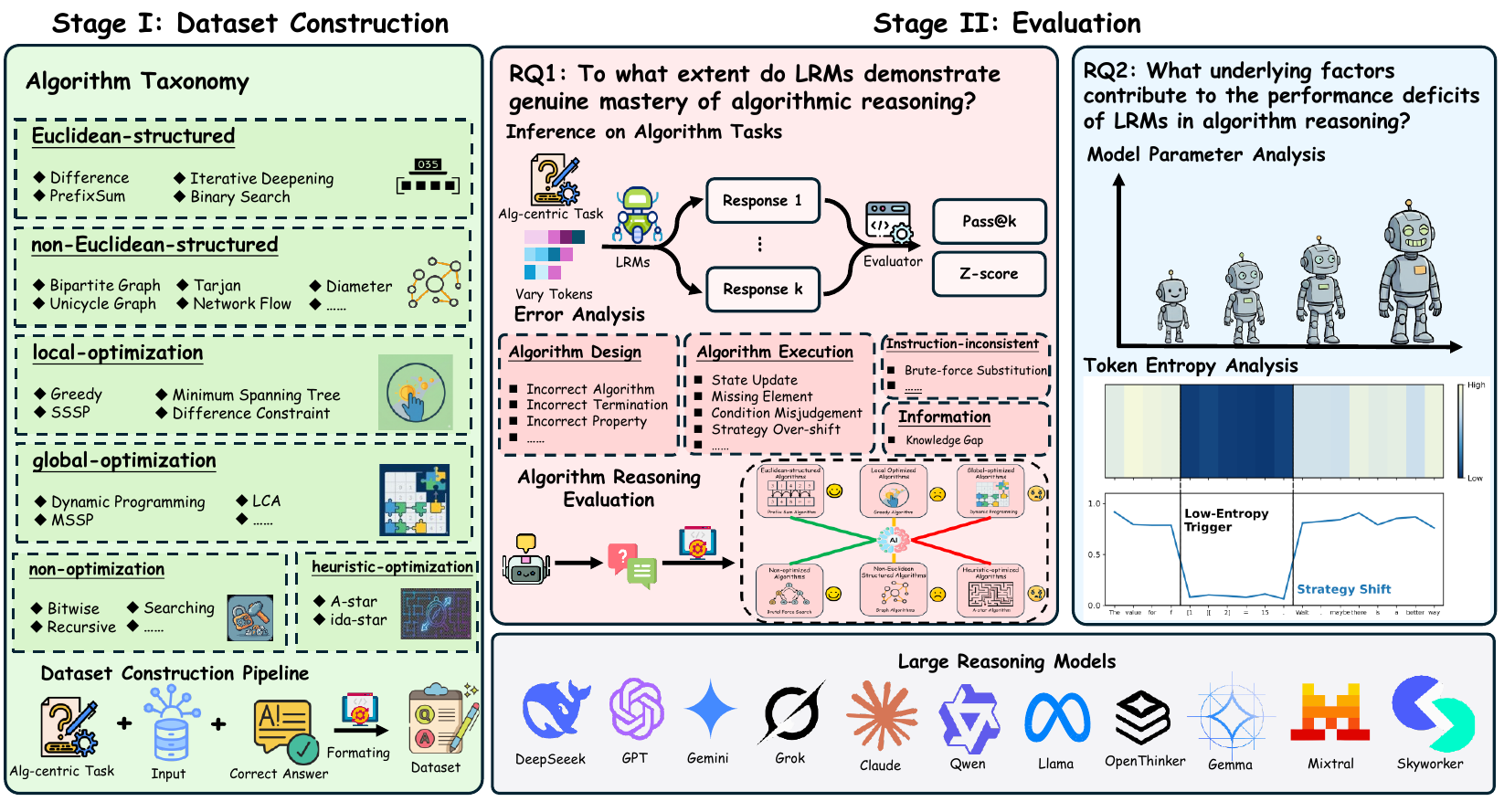}
  \caption{
    The overview of the proposed benchmark.
}
  \label{fig:benchmark_overview}
\end{figure*}

As discussed in Limitation 1 of Section~\ref{sec: intro}, existing algorithmic benchmarks predominantly draw their problem sets from open platforms, where individual problem often requires the simultaneous application of multiple algorithmic skills. This characteristic makes it difficult to accurately assess the performance of LLMs on specific and isolated algorithms. Such benchmark pipelines follow a \textbf{problem-centric paradigm}, emphasizing the models’ ability to solve problems rather than their capacity to comprehend and internalize the underlying algorithmic principles, as illustrated in Figure~\ref{fig:benchmark_overview}.

In contrast, our dataset construction follows an \textbf{algorithm-centric paradigm}. Human learning does not occur solely through exposure to diverse problems; instead, individuals first acquire an understanding of algorithms and their foundational ideas, and subsequently evaluate their mastery by applying these algorithms to relevant problems. Analogously, to enable LLMs to develop genuine algorithmic reasoning abilities, we argue that there is a pressing need for a new class of benchmarks—those designed from an algorithm-centric perspective—to evaluate model performance on individual algorithms in isolation.

It is important to note that classical algorithms are fundamentally devised to solve problems efficiently. This motivation has historically driven humans either to apply optimization techniques directly or to leverage data structures that improve the time–space complexity of computation. Consequently, the core ideas underlying classical algorithms can be broadly categorized into two groups, as illustrated in Figure~\ref{fig:benchmark_overview}: (1) Optimization-oriented ideas, encompassing non-optimized ideas (e.g., brute-force search), locally optimized ideas (e.g., greedy methods), globally optimized ideas (e.g., dynamic programming), and heuristic look-ahead idas (e.g., A-star); and (2) Structure-oriented ideas, which include both Euclidean-structured and non-Euclidean-structured algorithmic principles. A detailed explanation of these categories is provided in Appendix~\ref{appen: alg_taxonomy}.

\subsection{Dataset Curation}
\label{sec: dataset_curation}
The detailed statistics of our constructed dataset are presented in Table~\ref{tab: dataset_overview}. To address the challenges identified in \textit{Limitation 1} and \textit{Limitation 3} of Section~\ref{sec: intro}—specifically, the issues of inaccurate evaluation of LLMs on single-targeted algorithms and potential data contamination—we engaged a group of algorithm experts. All recruited experts are accomplished ACM competitors with extensive experience and deep understanding of complex algorithmic problem solving. Under their guidance, we manually curated a collection of 3,000 original problems covering 27 distinct algorithms, each designed to require exactly one specific algorithm and verified to have no overlap with any existing open-source platforms. This construction process effectively mitigates both the evaluation bias and data contamination concerns that hinder prior benchmarks. Furthermore, in order to evaluate the mastery of LLMs on a targeted algorithm with a fine-grained perspective, we divide each problem into three difficulty levels, according to the computation complexity.

\begin{table*}[htbp]
  \centering
  \caption{The detailed statistics of the AlgBench dataset.}
  \resizebox{\textwidth}{!}{
  \begin{tabular}{@{}cccccccccccccc@{}}
    \toprule
    Algorithm Taxonomy & \multicolumn{4}{c}{Euclidean-structured} & \multicolumn{7}{c}{Non-Euclidean-structured} & \multicolumn{2}{c}{Heuristic-optimized} \\
    \cmidrule(lr){1-1}
    \cmidrule(lr){2-5}
    \cmidrule(lr){6-12}
    \cmidrule(lr){13-14}
    Algorithms & ID & PS & Diff & BS & TDG & TUG & BGC & BGM & NF & DT & UG & AS & IDAS \\
    \midrule
    Problem Numbers & 30 & 60 & 60 & 150 & 120 & 120 & 120 & 120 & 120 & 120 & 120 & 120 & 120 \\
  \end{tabular}
  }
\resizebox{\textwidth}{!}{
  \begin{tabular}{@{}ccccccccccccccc@{}}
    \toprule
    Algorithm Taxonomy & \multicolumn{3}{c}{Non-optimized} & \multicolumn{4}{c}{Local-optimized} & \multicolumn{7}{c}{Global-optimized} \\
    \cmidrule(lr){1-1}
    \cmidrule(lr){2-4}
    \cmidrule(lr){5-8}
    \cmidrule(lr){9-15}
    Algorithms & BO & Recursion & Searching & Greedy & SSSP & MST & DC & LDP & IDP & TDP & BLDP & BDP & MSSP & LCA \\
    \midrule
    Problem Numbers & 30 & 150 & 120 & 120 & 120 & 120 & 120 & 150 & 150 & 120 & 150 & 150 & 120 & 120 \\
    \bottomrule
  \end{tabular}
  }
\label{tab: dataset_overview}
\end{table*}

\section{LLMs Evaluation}
\label{sec: eval}
In this section, we assess the efficacy of LRMs in solving diverse algorithmic problems and critically analyze the boundaries of their reasoning capabilities. Our evaluation addresses two primary research questions: \textbf{RQ1:} To what extent do LRMs demonstrate genuine mastery of algorithmic reasoning? \textbf{RQ2:} What underlying factors contribute to the performance deficits of LRMs in algorithms requiring global optimization strategies? The experimental setup and the evaluation settings are detailed in Appendix~\ref{appen: experimental_setup}. Meanwhile, due to the page limit, the additional experimental results are put in Appendix~\ref{appen: additional_expeiment}.

\subsection{Exploring Robust Proficiency of LRMs in Algorithm Reasoning (RQ1)}
\label{sec:RQ1}
While LRMs demonstrate generally high capabilities, their performance remains heterogeneous across distinct algorithmic domains. 

\noindent
\textbf{Evaluating LRMs' Ability towards Various Algorithmic Reasoning.} To provide a rigorous quantitative assessment of LRM mastery, we devised a set of problems designed to isolate individual algorithm. We stratified these problems according to the taxonomy introduced in Section~\ref{sec: alg_centric_paradigm} and subsequently evaluated LRM efficacy across these distinct categories. For each problem instance, we sampled $k$ responses from every comparison model and employed the Pass@$k$ metric for evaluation. Under this metric, a sample is deemed correct if at least one of the $k$ generated responses matches the ground-truth solution. Furthermore, to mitigate the impact of disparate task difficulties across the reasoning taxonomy, we adopted the normalization protocol from prior work~\cite{yu2023kola, yuan2024gracore}. Specifically, we computed the z-score of the Pass@$k$ accuracy for each model and task, subsequently linearly rescaling these scores to the interval [0, 1] as follows:

\begin{equation}
\begin{aligned}
z_{ij} &= \frac{x_{ij}-\mu(x_{i1},\cdots ,x_{i|M|})}{\sigma(x_{i1},\cdots ,x_{i|M|})}, \\
s_{ij} &= 1.0 \cdot \frac{z_{ij}-\operatorname{min}(z)}{\operatorname{max}(z)-\operatorname{min}(z)},
\end{aligned}
\end{equation}

\noindent
where $z_{ij}$ is the z-score of model $i$ on task $j$, and $s_{ij}$ is the normalized outcome.

\noindent
\textbf{Error Analysis.} We investigate the error typologies exhibited by LRMs when solving algorithmic problems. Our methodology relies on a defined taxonomy of error categories (refer to Appendix~\ref{appen: error_type}) implemented via an automated detection pipeline. Following prior work~\cite{guo2023gpt4graph, yuan2024gracore}, we first pre-process LRM outputs to enhance structural coherence. Specifically, we isolate responses containing incorrect terminal answers, segment them by double-newline delimiters, and index each resulting segment as $\langle i \rangle$. We then employ Qwen3-235B to aggregate these segments into semantic sections, summarizing each response into tuples of $(\text{start-index}, \text{end-index}, \text{summary})$. Finally, GPT-o3 annotates each defined section with one or more error categories, allowing us to compute the distribution of error types across the dataset.

\noindent
Due to the page limit, we present the main experimental results of strong LRMs in the main body, while putting the remaining in the Appendix~\ref{appen: additional_expeiment}. The resulting data, showed in Tables~\ref{tab: structured}, Figure~\ref{fig: radarchar_main} and Figure~\ref{fig: error_analysis}, yield the following key observations:

\begin{figure*}[htbp]
	\centering
  \includegraphics[width=\textwidth]{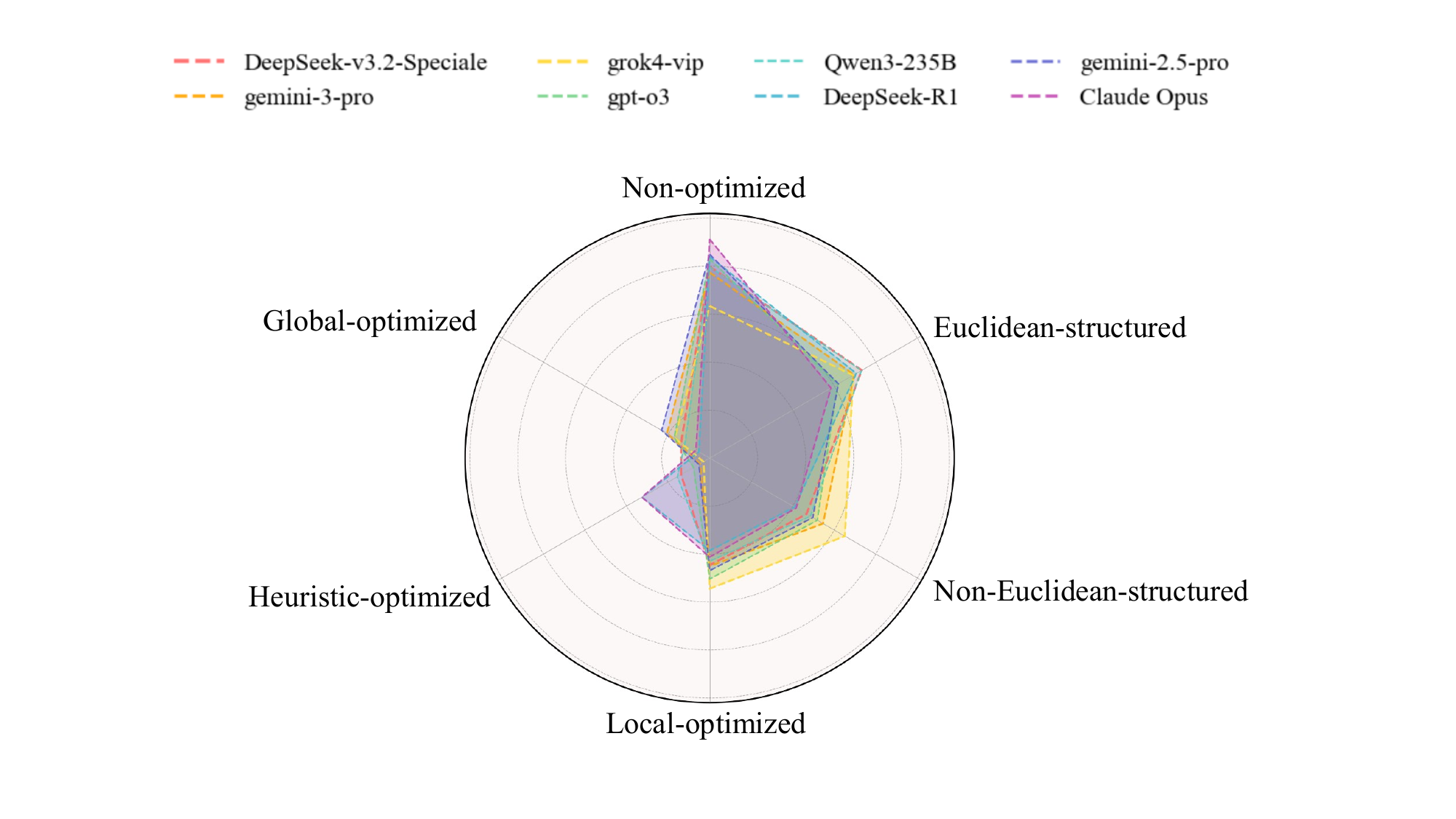}
  \caption{
    The normalized z-scores of LRMs on different algorithm taxonomies.
}
\label{fig: radarchar_main}
\end{figure*}

\noindent
\textbf{\ding{172} LRMs cannot effectively handle complex algorithm reasoning, such as global optimization and heuristic-optimization algorithm reasoning.} As shown in Tables~\ref{tab: structured} and Figure~\ref{fig: radarchar_main}, LRMs generally exhibit superior proficiency in Euclidean-structured and non-optimization algorithms. Conversely, their performance diminishes in non-Euclidean and local-optimization contexts, with the most significant deficits observed in global-optimization and heuristic-optimized algorithms. For instance, DeepSeek-v3.2-speciale attains 88\% and 92\% accuracy on Euclidean-structured and non-optimization algorithms, respectively. However, performance declines moderately to 70\% and 69\% on non-Euclidean and local-optimization tasks, followed by a precipitous drop to 49\% for both global-optimization and heuristic-optimization algorithms. This pattern of performance degradation was consistent across the majority of evaluated models.

The primary cause of this degradation pattern largely stems from the prevailing problem-centric paradigm adopted by most reinforcement learning (RL) fine-tuning strategies for LRMs. As discussed in Section \ref{sec: alg_centric_paradigm}, such a paradigm inherently fails to encode the fundamental ideas underlying algorithmic reasoning, thereby limiting LRMs’ ability to acquire and execute complex algorithm reasoning. Consider the case of Dynamic Programming (DP). Under the problem-centric approach, RL fine-tuning typically aggregates datasets containing DP problem instances, corresponding solutions, and reward signals, and then directly exposes the model to these data during training. However, this process provides no mechanism for the model to infer \textit{why} solutions to DP tasks take their specific structural form. It is well understood that a valid DP formulation must satisfy three key conditions: the presence of overlapping subproblems, the optimal substructure property, and the absence of aftereffects (i.e., Markovian structure). None of these conditions are explicitly manifested in the final solutions provided in the datasets. Consequently, LRMs are unable to reconstruct the appropriate DP states and transition functions required for systematic algorithm reasoning.

\begin{table*}[ht]
\centering
\caption{Leading LRMs' performance on problems with different algorimic ideas.}
\resizebox{0.7\textwidth}{!}{
\begin{tabular}{@{}ccccc>{\columncolor{red!10}}c|ccccccc>{\columncolor{red!10}}c@{}}
\toprule
\multirow{2}{*}{LLMs} & \multicolumn{5}{c|}{Euclidean-structured algorithms}                         & \multicolumn{8}{c}{non-Euclidean-structured algorithms}                                                                                                                               \\ \cmidrule(l){2-14} 
                                                & ID & PS & Diff & BS & Avg.     & TDG & TUG & BGC & BGM & NF & DT & UG & Avg.         \\ \midrule
DeepSeek-v3.2-Speciale  &  0.77  &  0.93  &  0.93  & 0.89  &  0.88  &  0.73  &  0.72  &  0.58  &  0.59  &  0.58  &  0.90  &  0.83  &  0.70  \\
gemini-3-pro & 0.93  & 0.93  & 0.48  & 0.95  & 0.83  & 0.86  & 0.58  & 0.56  & 0.62  & 0.63  & 0.90  & 0.78  & 0.70  \\
grok4-vip                                       &        0.80             &      0.78      &     0.96       &        0.66       &    0.80     &           0.87                &           0.72                  &             0.76                 &        0.70                  &      0.62        &     0.96              &       0.80          &     0.78        \\
gpt-o3                                          &        0.63             &     0.95       &     0.48       &       0.94        &    0.75     & 0.87                     & 0.75                          & 0.44                    & 0.58                     & 0.62        & 0.85                & 0.70              & 0.69       \\
Qwen3-235B                                  & 0.63               & 0.95         & 0.50         & 0.89         & 0.74  & 0.81                     & 0.50                          & 0.42                    & 0.56                    & 0.50           & 0.76             & 0.68            & 0.60 \\
DeepSeek-R1                                     &       0.64              &     0.93       &     0.50       &       0.91        &     0.74    & 0.59                     & 0.49                       & 0.50                    & 0.53                    & 0.52        & 0.75             & 0.65           & 0.58 \\
gemini-2.5-pro                                  &          0.63           &    0.88        &      0.50      &       0.55        &    0.64     & 0.67                     & 0.53                        & 0.37                    & 0.58                     & 0.48         & 0.80                & 0.48           & 0.56 \\
Claude Opus                                     &      0.64               &      0.57     &     0.50       &      0.71         &   0.60      & 0.67                     & 0.57                       &               0.31           & 0.56                    &     0.40         &        0.73           &     0.35            &     0.51        \\
gpt-oss-120B                                    & 0.40                  & 0.60         & 0.42      & 0.63         & 0.51  & 0.54                     & 0.42                       & 0.26                    & 0.40                       & 0.27        & 0.58              & 0.35              & 0.40       \\ 
\end{tabular}
}
\resizebox{0.7\textwidth}{!}{
\begin{tabular}{@{}ccccc>{\columncolor{red!10}}c|ccccccc>{\columncolor{red!10}}c@{}}
\toprule
\multirow{2}{*}{LLMs}        & \multicolumn{5}{c|}{local-optimized algorithms} & \multicolumn{8}{c}{global-optimized algorithms} \\ \cmidrule(lr){2-6} \cmidrule(l){7-14}
                             & Greedy & SSSP  & MST   & DC    & Avg.  & LDP   & IDP   & TDP   & BLDP  & BDP   & MSSP  & LCA   & Avg.  \\ \midrule
DeepSeek-v3.2-Speciale       & 0.51   & 0.78  & 0.69  & 0.78  & 0.69  & 0.63  & 0.65  & 0.18  & 0.54   & 0.50  & 0.38  & 0.53  & 0.49  \\
gemini-3-pro                 & 0.54   & 0.73  & 0.61  & 0.63  & 0.63  & 0.73  & 0.30  & 0.18  & 0.52  & 0.27  & 0.39  & 0.61  & 0.43  \\
grok4-vip                    & 0.56   & 0.82  & 0.69  & 0.77  & 0.71  & 0.47  & 0.62  & 0.17  & 0.47  & 0.54  & 0.48  & 0.60  & 0.48  \\
gpt-o3                       & 0.52   & 0.78  & 0.64  & 0.78  & 0.68  & 0.59  & 0.47  & 0.20  & 0.56  & 0.47  & 0.40  & 0.48  & 0.45  \\
Qwen3-235B               & 0.57   & 0.75  & 0.53  & 0.43  & 0.57  & 0.59  & 0.34  & 0.22  & 0.52  & 0.38  & 0.37  & 0.33  & 0.39  \\
DeepSeek-R1                  & 0.47   & 0.69  & 0.47  & 0.63  & 0.56  & 0.54  & 0.43  & 0.14  & 0.32  & 0.39  & 0.40  & 0.40  & 0.38  \\
gemini-2.5-pro               & 0.42   & 0.68  & 0.46  & 0.59  & 0.54  & 0.41  & 0.44  & 0.16  & 0.53  & 0.33  & 0.34  & 0.39  & 0.37  \\
Claude Opus                  & 0.48   & 0.68  & 0.30  & 0.58  & 0.51  & 0.26  & 0.39  & 0.19  & 0.43  & 0.26  & 0.37  & 0.32  & 0.32  \\
gpt-oss-120B                 & 0.36   & 0.62  & 0.50  & 0.61  & 0.52  & 0.30  & 0.29  & 0.12  & 0.41  & 0.23  & 0.21  & 0.39  & 0.28  \\ 
\end{tabular}}
\resizebox{0.7\textwidth}{!}{
\begin{tabular}{@{}cccc>{\columncolor{red!10}}c|cc>{\columncolor{red!10}}c@{}}
\toprule
\multirow{2}{*}{LLMs}        & \multicolumn{4}{c|}{non-optimized algorithms} & \multicolumn{3}{c}{heuristic-optimized algorithms} \\ \cmidrule(lr){2-5} \cmidrule(l){6-8}
                             & BO    & Recursion & Searching & Avg.  & AS    & IDAS    & Avg.  \\ \midrule
DeepSeek-v3.2-Speciale       & 0.97  & 0.93      & 0.87      & 0.92  & 0.39  & 0.58  & 0.49  \\
gemini-3-pro                 & 0.97  & 0.91      & 0.78      & 0.89  & 0.34  & 0.25  & 0.30  \\
grok4-vip                    & 0.90  & 0.53      & 0.87      & 0.77  & 0.40  & 0.39  & 0.39  \\
gpt-o3                       & 1.0   & 0.89      & 0.80      & 0.90  & 0.32  & 0.47  & 0.39  \\
Qwen3-235B               & 0.80  & 0.78      & 0.77      & 0.78  & 0.38  & 0.45  & 0.41  \\
DeepSeek-R1                  & 0.85  & 0.86      & 0.75      & 0.82  & 0.41  & 0.65  & 0.53  \\
gemini-2.5-pro               & 0.84  & 0.87      & 0.69      & 0.80  & 0.24  & 0.26  & 0.25  \\
Claude Opus                  & 0.85  & 0.89      & 0.61      & 0.79  & 0.37  & 0.55  & 0.46  \\
gpt-oss-120B                 & 0.47  & 0.37      & 0.60      & 0.48  & 0.26  & 0.30  & 0.28  \\  \bottomrule
\end{tabular}}
\label{tab: structured}
\end{table*}

\subsection{Exploring Underlying Mechanisms for Degradation Patterns of LRMs' algorithm Reasoning (\textbf{RQ2})}
\label{sec:RQ2}

In addressing \textbf{RQ1}, we observed that LRMs exhibit heterogeneous performance across distinct algorithmic reasoning tasks. Specifically, while the majority of LRMs demonstrate proficiency in solving non-optimized and Euclidean-structured problems, they show marked limitations when confronting global- and heuristic-optimized problems. In this section, we investigate the underlying mechanisms driving this performance disparity.

\noindent
\textbf{Model scaling analysis.} We begin by assessing the relationship between model's parameter and algorithmic reasoning performance. We utilize a comprehensive range of Qwen-series models for this analysis: Qwen3-4B, Qwen3-8B, Qwen3-14B, Qwen3-32B, and Qwen3-235B. Figure~\ref{fig: scaling_main} (see in Appendix~\ref{appen: additional_expeiment}) depicts the comparative results, detailing both the raw performance metrics and the incremental improvements observed over the base Qwen3-4B model.

\begin{figure*}[ht]
	\centering
  \includegraphics[width=0.75\textwidth]{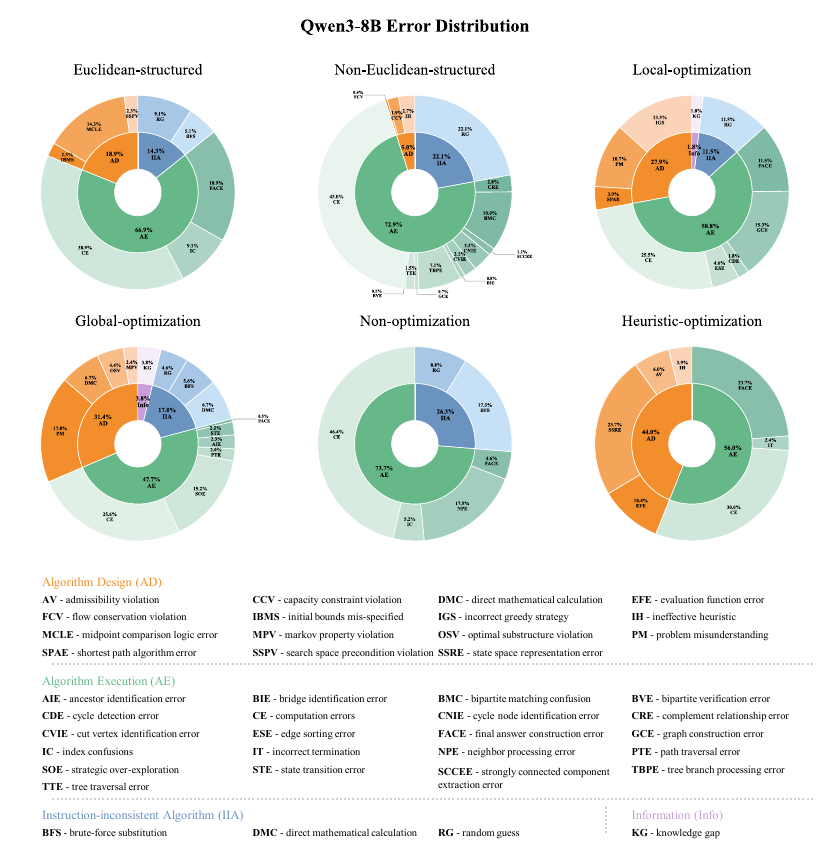}
  \caption{
    Error type distributions across algorithmic reasoning taxonomies for Qwen3-8B.
}
\label{fig: error_analysis}
\end{figure*}

\noindent
\textbf{Strategic over-exploration analysis.} Our error analysis reveals a relatively significant prevalence of strategic over-exploration within global-optimization algorithmic reasoning tasks, as depicted in Figure~\ref{fig: error_analysis}. To elucidate the mechanisms driving this phenomenon, we calculate the policy entropy of tokens proximal to strategic shift indicators (e.g., wait, but, maybe, and so on). The details of entropy calculation and the analysis process is illustrated in Appendix~\ref{appen: token_entropy_analysis}. The corresponding experimental results are reported from Figure~\ref{fig: qwen3_4b_token_entropy} to Figure~\ref{fig: qwen3_235B_token_entropy} in Appendix~\ref{appen: additional_expeiment}.

Observing the above experimental results, we summarize the findings as follows:

\noindent
\textbf{\ding{173} Improvements in the algorithmic reasoning capabilities of LRMs are not commensurate with the scaling of model parameters.} As illustrated in Figure~\ref{fig: scaling_main}, the benefits of model scaling are highly non-uniform across algorithm taxonomies. LRMs exhibit substantial performance gains in local-optimized and non-Euclidean-structured reasoning, whereas improvements in global- and heuristic-optimized tasks are marginal. Notably, scaling yields negligible benefits for non-optimized and Euclidean-structured problems. For instance, the Qwen-series achieves relative improvements exceeding 120\% and 80\% in the local-optimized and non-Euclidean categories, respectively, compared to gains of less than 20\% in non-optimized and Euclidean-structured domains.

Recall Finding 1 from Section~\ref{sec:RQ1}, which established that LRMs exhibit superior performance in Euclidean-structured and non-optimized algorithms, but only moderate performance in local-optimized and non-Euclidean-structured contexts. Consequently, the limited scaling gains observed in the former categories can be attributed to \textbf{performance saturation}: since LRMs have already achieved high proficiency in these areas, the margin for further improvement is constrained. Conversely, non-Euclidean-structured and local-optimized algorithms are characterized by limited observation-overhead (i.e., moderate algorithmic difficulty). In these domains, the complexity is sufficient to benefit from increased capacity, which means scaling laws remain highly effective in these two algorithmic reasoning.

However, improvements in global- and heuristic-optimized algorithmic reasoning remain marginal, as well as these domains exhibiting the lowest absolute performance (see Table~\ref{tab: structured}, Figure~\ref{fig: radarchar_main}, and Figure~\ref{fig: scaling_main}). This stagnation corroborates our observation in Section~\ref{sec:RQ1} that the prevailing problem-centric RL paradigm is insufficient for acquiring abstract algorithmic ideas. Consequently, we advocate for a paradigm shift from problem-centric to algorithm-centric RL. This transition is essential to effectively leverage scaling laws and develop LRMs with robust algorithmic reasoning capabilities.

\begin{problembox}{Example 1}
Let me think about this problem. We need to track both the total D and P, and the number of contestants selected, perhaps the state can be represented as $\operatorname{dp}[i][k][d][p]$, and the transition function can be $\operatorname{dp}[i][k][d][p]=\max \left \{ \operatorname{dp}[i-1][j][m-D_j][n-P_j]\right \}$. Thus, the initial state is $\operatorname{dp}[0][0][0][0]=1$, $\operatorname{dp}[1][0][0][0]=0$, $\cdots$. \textcolor{red}{Wait, maybe there is a better solution}. Let me rethink about how to model this. $\cdots$  
\end{problembox}


\noindent
\textbf{\ding{174} The presence of necessary low-entropy tokens hinders the successful completion of algorithmic reasoning processes, thereby inducing multiple strategic over-shifts.} Recall that the error analysis results presented in Figure~\ref{fig: error_analysis} indicate a relatively high prevalence of strategic over-exploration in global-optimized algorithmic reasoning, a challenge that most LRMs struggle to address effectively. A closer examination of the false responses reveals an intriguing phenomenon: although LRMs are capable of constructing correct algorithms, they often fail to faithfully execute the designed procedures and instead prematurely abandon them in favor of alternative strategies, as shown in the Example 1. 

Considering the maximum-entropy principle underlying prevailing RL paradigms, we analyze the distribution of low-entropy tokens in the vicinity of indicator tokens, as illustrated from Figure~\ref{fig: qwen3_4b_token_entropy} to Figure~\ref{fig: qwen3_235B_token_entropy}, with a detailed illustration about token entropy calculation and analysis illustrated in Appendix~\ref{appen: token_entropy_analysis}. The results substantiate our finding that the presence of necessary low-entropy tokens in globally optimized algorithms (e.g., numerical constants, delimiters such as “][”, and related structural symbols) can impede the successful completion of global optimization–oriented algorithmic reasoning. This observation stands in tension with the dominant maximum-entropy RL paradigm, suggesting that not all low-entropy tokens should be uniformly penalized during algorithmic reasoning. Instead, RL fine-tuning could explicitly assign higher rewards to such indispensable low-entropy tokens, or alternatively, adopt agent-style paradigms that incorporate auxiliary tools to assist LRMs in executing reasoning steps when this phenomenon arises.


\section{Conclusion}
\label{sec: conclusion}
In this study, we presented AlgBench, a benchmark designed to move beyond problem-centric evaluation toward an algorithm-centric understanding of LRM capabilities. Through the manual curation of 3,000 original problems by algorithmic experts, we provided a contamination-free benchmark to rigorously test LRMs' mastery over specific algorithmic taxonomies.

Our comprehensive evaluation yielded three critical insights into the current state of algorithmic reasoning in LRMs:
\begin{itemize}
    \item \textbf{Performance Heterogeneity:} LRMs exhibit superior proficiency in Euclidean-structured and non-optimized algorithms but struggle significantly with global-optimization and heuristic-optimization tasks.

    \item \textbf{Scaling Limitations:} Improvements in algorithmic reasoning are not commensurate with model parameter scaling. While scaling benefits algorithms like local-optimization and non-Euclidean-structured ones, it yields marginal gains for complex global- and heuristic optimization domains, suggesting that performance in these areas has reached a plateau under current training paradigms.

    \item \textbf{Strategic Over-shifts and Low-Entropy Barriers:} We identified that the presence of indispensable low-entropy tokens (e.g., structural delimiters and numerical constants) often triggers "strategic over-shifts". Models may construct correct algorithmic states but fail during execution, prematurely abandoning valid strategies in favor of alternative, often incorrect, reasoning paths.
\end{itemize}

Ultimately, our results substantiate that a paradigm shift from problem-centric to algorithm-centric RL is essential for LRMs to achieve robust, human-level algorithmic reasoning.

\section*{Limitations}
Although the proposed AlgBench provides an effective evaluation of the algorithmic reasoning capabilities of leading LRMs, it nevertheless exhibits several limitations, which are discussed from three perspectives below.

\noindent
\textbf{Complex Competitive Algorithm Gap.} Although our dataset covers 27 algorithms, several complex competitive algorithms are not included in the current version (e.g., Segment Tree, Aho–Corasick and so on). Expanding the benchmark to incorporate such algorithms remains an important direction for future work.

\noindent
\textbf{Ambiguous Difficulty Quantification of Algorithm Taxonomies.} Although our work proposes a novel algorithmic taxonomy comprising six fundamental algorithmic principles, it remains challenging to rigorously quantify the relative difficulty of these algorithms and to establish a clear relationship between algorithmic difficulty and the algorithmic reasoning capabilities of LRMs.

\noindent
\textbf{Limited Dataset Size.} Although our dataset comprises over 3,000 problems for evaluating the algorithmic reasoning capabilities of LRMs, it remains insufficient for pre-training or fine-tuning LRMs. Consequently, continually expanding the dataset is essential to support more robust and effective pre-training or fine-tuning in future work.


\bibliography{reference}

\appendix

\section{Experimental Setup}
\label{appen: experimental_setup}
\noindent
\textbf{Evaluated LRMs.} Our evaluation encompasses 25 state-of-the-art LRMs, including both closed-source commercial models and open-source LRMs. Leveraging AlgBench, we conduct a comprehensive assessment of the algorithmic reasoning capabilities of contemporary LRMs. The evaluated models are listed in Table~\ref{tab:eval_LRMs}:

\begin{table}[htbp]
    \centering
    \caption{The detailed information about evaluated LRMs.}
\resizebox{0.45\textwidth}{!}{
    \begin{tabular}{@{}ccc@{}}
\toprule
Model Name & Organization & Release Date \\
\midrule
DeepSeek-v3.2-Speciale & DeepSeek & December 2025 \\
gemini-3-pro & Google DeepMind & November 2025 \\
grok4 & xAI & July 2025 \\
gpt-o3 & OpenAI & April 2025 \\
Qwen3-235B & Alibaba & July 2025 \\
DeepSeek-R1 & DeepSeek & January 2025 \\
gemini-2.5-pro & Google DeepMind & April 2025 \\
Claude Opus & Anthropic & November 2025 \\
gpt-oss-120B & OpenAI & August 2025 \\ \midrule
Qwen3-4B & Alibaba & April 2025 \\
Qwen3-8B & Alibaba & April 2025 \\
Qwen3-14B & Alibaba & April 2025 \\
Qwen3-32B & Alibaba & April 2025 \\
QwQ-32B & Alibaba & March 2025 \\
DeepSeek-R1-0528-Qwen3-8B & DeepSeek & May 2025 \\
DeepSeek-R1-Distill-Qwen-32B & DeepSeek & May 2025 \\
OpenThinker2-32B & OpenThoughts & March 2025 \\
Skyworker-OR1-32B & SkyworkAI & May 2025 \\
Qwen3-Coder-30B-A3B-Instruct & Alibaba & July 2025 \\
Llama-3.3-70b-instruct & Meta & December 2024 \\
gemma3-12B & Google & March 2025 \\
gemma3-27B & Google & March 2025 \\
Llama3-8B & Meta & September 2024 \\
Llama3.1-8B & Meta & September 2024 \\
Mixtral-8x7B & Mistral AI & December 2023 \\
\bottomrule
\end{tabular}}
    \label{tab:eval_LRMs}
\end{table}

\noindent
\textbf{Evaluation Settings.} We employ pass@$k$ to evaluate the average performance of LRMs on algorithmic reasoning, and the z-score to mitigate the impact of disparate problems' difficulties across the algorithmic reasoning taxonomy, as illustrated in Section~\ref{sec:RQ1}. To evaluate the performance of LRMs on a target algorithm, each prompt is appended with explicit instructions (e.g., “Please solve this problem using the XXX algorithm”). The detailed prompt templates are provided in Appendix~\ref{appen: aaqe}.

To maintain consistency in evaluations and facilitate reproduction, we set the maximum output length to 32k tokens and employ a greedy decoding strategy with temperature 0.6.

\noindent
\textbf{Experiment Environment.} To ensure reproducibility, we report the hardware and software configurations employed in our experiments. All experiments were conducted on a server equipped with an Intel(R) Xeon(R) Gold 6240 CPU (2.60 GHz) and 8 NVIDIA A800 GPUs, each with 80 GB of memory, using CUDA version 12.4.

\section{Related Work}
\label{appen: related_work}
\noindent
\textbf{Benchmarking LRMs on Math Problems.} Mathematical reasoning has long attracted many attentions from both academic and industry. Multiple math-related benchmarks have been proposed in this area~\cite{koncel2016mawps, hendrycks2021measuring, cobbe2021training, chen2022program, he2024olympiadbench, xu2025ugmathbench, petrov2025proof}. Among these, MATH-500~\cite{hendrycks2021measuring}, AIME 2024~\cite{he2024olympiadbench}, and AIME 2025~\cite{petrov2025proof} are those among the most representative datasets for the training and evaluation of LRMs on high-level mathematical reasoning abilities. However, as stated in Section~\ref{sec: intro}, these training and evaluation cannot help improve or testify the reasoning capacity of LRMs on algorithm reasoning, since there are significant difference between these two reasoning domains. Thus, in order to thoroughly promote the development of LRMs towards genuine artificial general intelligence (AGI), it is necessary for building benchmarks for LRMs' algorithmic reasoning abilities as well.

\noindent
\textbf{Benchmarking LRMs on Competitive Algorithmic Programming.} As algorithmic reasoning constitutes a fundamental capability that LRMs must acquire to advance toward genuine AGI, a number of benchmarks have been proposed to evaluate such abilities~\cite{hendrycks2021measuringcode, li2022competition, li2023taco, xia2025leetcodedataset, white2024livebench, jain2024livecodebench, zheng2025livecodebench}. Among these, several benchmarks achieve considerable influence among academic and industrial areas, such as LeetCodeDataset~\cite{xia2025leetcodedataset}, LiveCodeBench~\cite{jain2024livecodebench} and LiveCodeBench Pro~\cite{zheng2025livecodebench}. However, all of these existing benchmarks predominantly adopt a problem-centric paradigm, emphasizing the accuracy of LRMs in solving algorithmic tasks rather than assessing whether they have genuinely internalized the underlying principles of a specific algorithm. To achieve this objective, and to the best of our knowledge, we are the first to introduce a paradigm shift toward an algorithm-centric framework, which places primary emphasis on assessing whether LRMs have genuinely mastered an algorithm and its underlying core principles.

\section{Definition and Explanation}
\label{appen: D&E}

\subsection{Detailed Illustration of Algorithm Taxonomy}
\label{appen: alg_taxonomy}
This appendix provides a detailed rationale for the taxonomy used in the dataset. The algorithms are categorized based on the \textbf{algorithm-centric paradigm}.

\subsubsection{Euclidean-structured}

\paragraph{Category Characteristics:}
This category encompasses algorithms that operate on data structures with a defined linear order or coordinate system (e.g., arrays, sequences, grids). The core feature is that the data elements have fixed, indexable positions. Problem-solving relies heavily on dimension partitioning, interval operations, or searching within a linear metric space.

\paragraph{Algorithm Justification:}
\begin{description}[style=standard, leftmargin=1.5em, itemsep=0.5em, 
]
    \item[Difference Arrays \& Prefix Sum:] Both algorithms strictly depend on the continuity and ordering of linear arrays. They utilize the sequential arrangement of elements to perform range updates or queries in $O(1)$ time, leveraging the Euclidean property of distance and position.
    
    \item[Binary Search:] This algorithm is predicated on the "sorted" property, which implies a strong linear constraint. It operates in a 1D Euclidean space (a number line), systematically reducing the search interval based on coordinate comparison.
    
    \item[Iterative Deepening:] While often applied to graphs, the mechanism itself maps a complex search space onto a linear dimension (depth $d=1, 2, 3 \dots$). It treats the search radius as a gradually expanding Euclidean boundary.
\end{description}

\subsubsection{Non-Euclidean-structured}

\paragraph{Category Characteristics:} 
Algorithms in this category deal with topological structures where elements (nodes) are connected by relations (edges) rather than fixed coordinates. The focus is on connectivity, structural shape (cycles, components), and flow, rather than spatial position.

\paragraph{Algorithm Justification:}
\begin{description}[style=standard, leftmargin=1.5em, itemsep=0.5em, 
]
    \item[Bipartite Graph, Diameter, \& Unicyclic:] These tasks involve analyzing the intrinsic shape of the graph. Identifying a bipartite set, finding the longest path (diameter), or detecting a ring in a tree (unicyclic) are purely topological properties independent of embedding.
    \item[Connectivity (Directed/Undirected):] Connectivity algorithms (finding SCCs or BCCs) analyze the reachability among nodes, a fundamental property of non-Euclidean graph theory.
    \item[Network Flow \& Bipartite Matching:] These are complex graph problems that view edges as capacities or relationships. They optimize flow or associations based strictly on the graph's adjacency matrix, not on geometric proximity.
\end{description}

\subsubsection{Non-optimized}

\paragraph{Category Characteristics:} 
These algorithms are fundamentally procedural or computational in nature. Rather than incorporating optimization principles to reduce time or space complexity, they rely on deterministic computations, logical operations, or exhaustive traversal strategies to produce correct outputs.

\paragraph{Algorithm Justification:}
\begin{description}
    \item[Bitwise:] Involves low-level arithmetic and logical manipulation of binary representations. It is a calculation process, not an optimization search.
    \item[Recursive:] Represents a control flow mechanism for task decomposition. Without an associated cost function or memoization, pure recursion is simply a method of execution.
    \item[Searching (BFS/DFS):] Fundamental traversal strategies. Used alone, the goal is visiting nodes or verifying existence, not optimizing a numerical value.
\end{description}

\subsubsection{Local-optimized}

\paragraph{Category Characteristics:} 
This category encompasses algorithms that iteratively construct solutions by selecting locally optimal actions at each step, such as those satisfying the greedy-choice property, or by applying local relaxation operations. Such algorithms generally do not preserve a comprehensive global state history and are predicated on the assumption that a sequence of local improvements will converge to a globally optimal solution.

\paragraph{Algorithm Justification:}
\begin{description}
    \item[Greedy:] Greedy algorithm is one of the most classical local-optimized algorithms. It makes irrevocable decisions only based on locally optimal strategies without observing the global state information.
    \item[Shortest Single (Dijkstra/Bellman-Ford):] These algorithms are based on relaxation, a local operation that evaluates whether a candidate path can be improved through a given edge. In particular, Dijkstra’s algorithm adopts a greedy strategy by iteratively selecting the unvisited node with the minimum tentative distance.
    \item[MST (Prim/Kruskal):] The construction of a Minimum Spanning Tree is a canonical example of a greedy paradigm. Algorithms such as Prim’s and Kruskal’s incrementally expand the spanning structure by selecting, respectively, the nearest admissible node or the minimum-weight edge. These selection rules are inherently locally optimal, as they rely exclusively on immediate, local information rather than on a globally optimized view of the graph.
    \item[Difference Constraints:] These systems are typically solved by transforming them into Single-Source Shortest Path problems, thus inheriting the local relaxation characteristics of Bellman-Ford or SPFA.
\end{description}

\subsubsection{Global-optimization}
\paragraph{Category Characteristics:} 
Algorithms in this category address problems by decomposing them into overlapping subproblems and systematically integrating their solutions. A defining characteristic is the maintenance of a global state—typically through dynamic programming tables or memoization—which guarantees that the final solution is globally optimal by accounting for all relevant state transitions.

\paragraph{Algorithm Justification:}
\begin{description}
    \item[DP Variants] \textbf{(Linear, Interval, Tree, Bitmask and Binarly Lifting):} Dynamic Programming is the epitome of global optimization. It systematically explores all relevant states and transitions, using the valid DP transition function to guarantee the global optimum is found from sub-solutions.
    \item[LCA:] Although primarily employed to answer queries, the preprocessing stage constructs a global state table (e.g., $ancestor[i][j]$, denoting the $2^j$-th ancestor of node $i$). This procedure is governed by the state transition relation $f(i, j) = f\left(f(i, j-1), j-1\right)$, which exemplifies the global state propagation and optimization logic characteristic of dynamic programming.
    \item[Shortest Multi (Floyd-Warshall):] In contrast to Dijkstra’s algorithm, Floyd–Warshall adopts a dynamic programming framework that systematically accounts for the effect of every node as a potential intermediate vertex between all pairs of nodes, thereby achieving a globally optimized solution over the entire distance matrix.
\end{description}

\subsubsection{Heuristic-optimization}

\paragraph{Category Characteristics:} 
This category serves as an intermediate paradigm between deterministic optimization methods and heuristic search strategies. It incorporates domain-specific knowledge through a heuristic evaluation function $f(n)=g(n)+h(n)$, which combines the accumulated cost of the current state with an estimate of the remaining cost to the goal. By jointly accounting for present progress and future potential, this formulation guides and prioritizes the search process, enabling more efficient solution discovery.

\paragraph{Algorithm Justification:}
\begin{description}
    \item[A* (A-Star):] One of the most classical algorithms for heuristic search. It optimizes the path-finding process by adding an estimated future cost $h(n)$ to the known path cost, guiding the search intelligently.
    \item[IDA*:] Combines the space efficiency of Iterative Deepening with the directional guidance of heuristics. It uses the heuristic threshold to prune branches that exceed the estimated optimal cost.
\end{description}

\subsection{Error Type Classification}
\label{appen: error_type}
We categorize model errors into four major classes:
\subsubsection{Algorithm Design (AD)}Errors in this category arise when the model fails to formulate a theoretically sound algorithmic approach for the given problem.
\begin{description}
    \item[\textbf{AV} (Admissibility Violation)] The heuristic function used in search algorithms (e.g., A*) overestimates the true cost to the goal ($h(n) > \text{actual\_cost}$), invalidating optimality guarantees or causing the pruning of optimal paths.
    \item[\textbf{CCV} (Capacity Constraint Violation)] In network flow problems, the model proposes a flow assignment that exceeds edge capacities or results in invalid negative flows.
    \item[\textbf{DMC} (Direct Mathematical Calculation)] The model attempts to bypass the required algorithmic design (e.g., Dynamic Programming) by deriving the answer directly through a mathematical formula or shortcut.
    \item[\textbf{EFE} (Evaluation Function Error)] The total estimated cost $f(n)$ is calculated incorrectly (e.g., $f(n) \neq g(n) + h(n)$) within heuristic search algorithms.
    \item[\textbf{FCV} (Flow Conservation Violation)] The model violates flow conservation constraints (inflow $\neq$ outflow) for non-source/sink nodes in network flow problems.
    \item[\textbf{IBMS} (Initial Bounds Mis-specified)] In binary search, the starting bounds (low/high) fail to cover the feasible range or exclude the correct answer.
    \item[\textbf{IA} (Incorrect Algorithm)] The model selects an algorithm that is fundamentally unsuitable for the problem constraints or objectives (e.g., using a greedy approach for a problem requiring DP).
    \item[\textbf{IGS} (Incorrect Greedy Strategy)] The model identifies a flawed local criterion for making greedy choices, leading to suboptimal or invalid solutions, often failing the exchange argument.
    \item[\textbf{IH} (Ineffective Heuristic)] The heuristic is technically admissible (e.g., $h(n)=0$) but provides poor guidance, causing the search to degrade into a less efficient algorithm (e.g., A* becoming Dijkstra's)12.
    \item[\textbf{MCLE} (Midpoint Comparison Logic Error)] The comparison logic in binary search is inverted or incorrect, leading to search directions that exclude the target.
    \item[\textbf{MPV} (Markov Property Violation)] In Dynamic Programming, the state definition fails to encapsulate all necessary history, leading to invalid transitions.
    \item[\textbf{OSV} (Optimal Substructure Violation)] The model fails to recognize that the optimal solution can be constructed from optimal sub-solutions, leading to incorrect DP state definitions or recurrence relations.
    \item[\textbf{PM} (Problem Misunderstanding)] The model misinterprets the problem's constraints, objective function, or edge cases, resulting in a flawed algorithmic model.
    \item[\textbf{SPAE} (Shortest Path Algorithm Error)] In difference constraints or graph problems, the model selects an inappropriate shortest path algorithm (e.g., using an algorithm that cannot handle negative weights when they are present).
    \item[\textbf{SSPV} (Search Space Precondition Violation)] The model applies an algorithm (e.g., binary search) without verifying essential preconditions, such as array sortedness or monotonicity.
    \item[\textbf{SSRE} (State Space Representation Error)] The definition of the search space is flawed, including incomplete state definitions, incorrect successor generation, or miscalculated edge costs.
\end{description}

\subsubsection{Algorithm Execution (AE)}

Errors in this category occur when the high-level logic is generally correct, but mistakes are made during the step-by-step implementation or calculation of the algorithm.

\begin{description}

\item[\textbf{AIE} (Ancestor Identification Error)] In Lowest Common Ancestor (LCA) problems, the model fails to correctly identify the common ancestor or finds one that is not the lowest.

\item[\textbf{APE} (Augmenting Path Error)] The model fails to correctly identify or construct augmenting paths in matching or flow problems, or terminates the search prematurely.

\item[\textbf{BIE} (Bridge Identification Error)] In graph connectivity, the model uses incorrect conditions (e.g., comparing $low[child]$ and $dfn[current]$) to identify bridge edges.

\item[\textbf{BMC} (Bipartite Matching Confusion)] The model misapplies theorems (e.g., K"{o}nig's theorem) regarding the relationship between vertex covers, independent sets, and matchings.

\item[\textbf{BPE} (Bipartite Partition Error)] The model incorrectly assigns vertices to partitions or fails to identify the bipartite structure of the graph.

\item[\textbf{BVE} (Bipartite Verification Error)] The model incorrectly determines graph bipartiteness or assumes it without verification.

\item[\textbf{CDE} (Cycle Detection Error)] The model fails to detect cycles during graph construction (e.g., in MST algorithms), allowing invalid edges to be added.

\item[\textbf{CE} (Computation Errors)] General arithmetic mistakes occurring during intermediate calculations, prefix sum accumulations, or variable updates.

\item[\textbf{CNIE} (Cycle Node Identification Error)] In unicyclic graphs, the model confuses nodes belonging to the cycle with those in attached tree branches.

\item[\textbf{CRE} (Complement Relationship Error)] The model fails to correctly apply the complement relationship (e.g., $|Max Independent Set| = |V| - |Min Vertex Cover|$).

\item[\textbf{CVIE} (Cut Vertex Identification Error)] The model incorrectly identifies articulation points, often mishandling root node cases or low-link value conditions.

\item[\textbf{ESE} (Edge Sorting Error)] In algorithms like Kruskal's, the model fails to sort edges correctly by weight.

\item[\textbf{FACE} (Final Answer Construction Error)] The algorithmic execution is sound, but the model fails to format, index, or construct the final output correctly.

\item[\textbf{GCE} (Graph Construction Error)] The model makes mistakes in building the underlying graph structure, such as missing edges or incorrect node representations.

\item[\textbf{IC} (Index Confusions)] The model confuses 0-based and 1-based indexing, or mixes row/column indices, leading to off-by-one errors.

\item[\textbf{IT} (Incorrect Termination)] The algorithm terminates at the wrong moment (e.g., A* stopping when a goal is seen rather than popped from the priority queue).

\item[\textbf{NPE} (Neighbor Processing Error)] The model incorrectly identifies valid neighbors, fails boundary checks, or processes neighbors in an invalid order during graph traversal.

\item[\textbf{PQE} (Priority Queue Error)] The model mismanages priority queue operations, such as incorrect priority updates or extraction logic.

\item[\textbf{PTE} (Path Traversal Error)] The model fails to correctly navigate paths in trees or graphs, missing nodes or making wrong directional decisions.

\item[\textbf{ROE} (Relaxation Operation Error)] In shortest path algorithms, the model fails to correctly implement the edge relaxation step or update distances.

\item[\textbf{SCCEE} (SCC Extraction Error)] In Tarjan's algorithm, the model fails to correctly identify the condition for extracting an SCC from the stack, where SCC stands for Strongly Connected Component.
\item[\textbf{SOE} (Strategic Over-exploration)] In Dynamic Programming, the model establishes correct states but wastes computation searching for unnecessary "better" strategies instead of executing the transition.
\item[\textbf{STE} (State Transition Error)] The model applies the DP transition formula incorrectly, using wrong indices or logic during the update step.
\item[\textbf{TBPE} (Tree Branch Processing Error)] In unicyclic graphs, the model mishandles the processing of tree components attached to cycle nodes.
\item[\textbf{TTE} (Tree Traversal Error)] The model implements incorrect logic for DFS/BFS traversals on trees, such as visiting nodes redundantly.
\end{description}

\subsubsection{Instruction-inconsistent Algorithm (IIA)}
Errors where the model deviates from the specific algorithmic method requested in the prompt, regardless of the correctness of the alternative solution.
\begin{description} 
    \item[\textbf{BFS} (Brute-force Substitution)] The model resorts to exhaustive enumeration or nested loops instead of the requested efficient algorithm (e.g., replacing DP or binary search with linear scans).
    \item[\textbf{DMC} (Direct Mathematical Calculation)] The model skips the requested algorithmic process (e.g., DP or simulation) entirely in favor of a direct formula derivation.
    \item[\textbf{EU} (Execution Unwillingness)] The model explicitly refuses to perform the detailed steps of the algorithm (e.g., refusing to compute a prefix sum matrix).
    \item[\textbf{RG} (Random Guess)] The model abandons logical reasoning and provides a baseless answer or switches to an unrelated method without justification. 
\end{description}

\subsubsection{Information (Info)} 
\begin{description} 
    \item[\textbf{KG} (Knowledge Gap)] The model demonstrates a fundamental lack of understanding of key concepts (e.g., optimal substructure, Markov property, or specific graph theorems) required to solve the problem.
\end{description}

\subsection{Detailed Illustration of Token Entropy Analysis}
\label{appen: token_entropy_analysis}
\noindent
\textbf{Token Entropy Calculation.} We follow the prior work of entropy calculation and obtain the corresponding token entropy in our experiments~\cite{shannon1948mathematical, kuhn2023semantic, wang2025beyond, xu2025tecp}. Formally, given the policy distribution $p_i$ at token index $i$ and the vocabulary $V$, the corresponding token entropy $H(p_i)$ is defined as follow:

\begin{equation}
\begin{aligned}
    H(p_i) = -\sum_{j=0}^{|V|} p_i(v_j)\operatorname{log}p_i(v_j). \\
\end{aligned}
\end{equation}

\noindent
\textbf{Token Entropy Analysis.} After computing token entropy, we rank all tokens in each LRM-generated response according to their entropy values and identify the corresponding top-$k$ high-entropy tokens and bottom-$k$ low-entropy tokens. We then quantify the co-occurrence between indicator tokens (e.g., wait, but, and so on) and the selected high-/low-entropy tokens by counting their occurrences within a predefined local context window around each indicator token (specifically, a distance of 10 tokens in our experiments). Based on this procedure, we obtain the experimental results presented in Figures~\ref{fig: qwen3_4b_token_entropy} through \ref{fig: qwen3_235B_token_entropy}. 

\subsection{Additional Algorithm Question Explanation}
\label{appen: aaqe}
This section illustrates several examples of our prompts about the algorithm questions, including the system prompts and the question prompts, as shown in the Example 2 and 3 below.

\begin{problembox}{Example 2: System Prompt}
You are a professional algorithm assistant. Please answer questions according to the following requirements:

1. Do not use any code execution or programming tools to calculate the answer.

2. You must give your thinking process before outputting the final result in the format "Final answer: [number1, number2, number3, ...]".

3. Make sure the output array format is correct, with numbers separated by commas and spaces.

\textcolor{red}{Note: Do not rely on code execution results}.
\end{problembox}

\begin{problembox}{Example 3: Question Prompt}
There are 8 students ready to take photos, arranged in 3 rows with [6, 1, 1] persons respectively. The first row is in the front, and the last row is in the back. The students have distinct heights and are labeled from 1 to 8 in descending order of height. During the photo arrangement, it is required that in each row, the heights decrease from left to right, and in each column, the heights decrease from back to front. How many different valid arrangements are possible for the group photo? \textcolor{red}{Please solve this using a linear dynamic programming algorithm}.
\end{problembox}

As illustrated in Section~\ref{sec: intro} and Appendix~\ref{appen: experimental_setup}, one of the primary goals of this benchmark is to evaluate LRMs' algorithmic reasoning ability towards a targeted algorithm. Thus, we specifically require LRMs not to use any code execution in system prompt and also require LRMs to use the specified algorithm in the question prompt.

\section{Additional Experimental Results}
\label{appen: additional_expeiment}
\begin{figure*}[htbp]
	\centering
  \includegraphics[width=\textwidth]{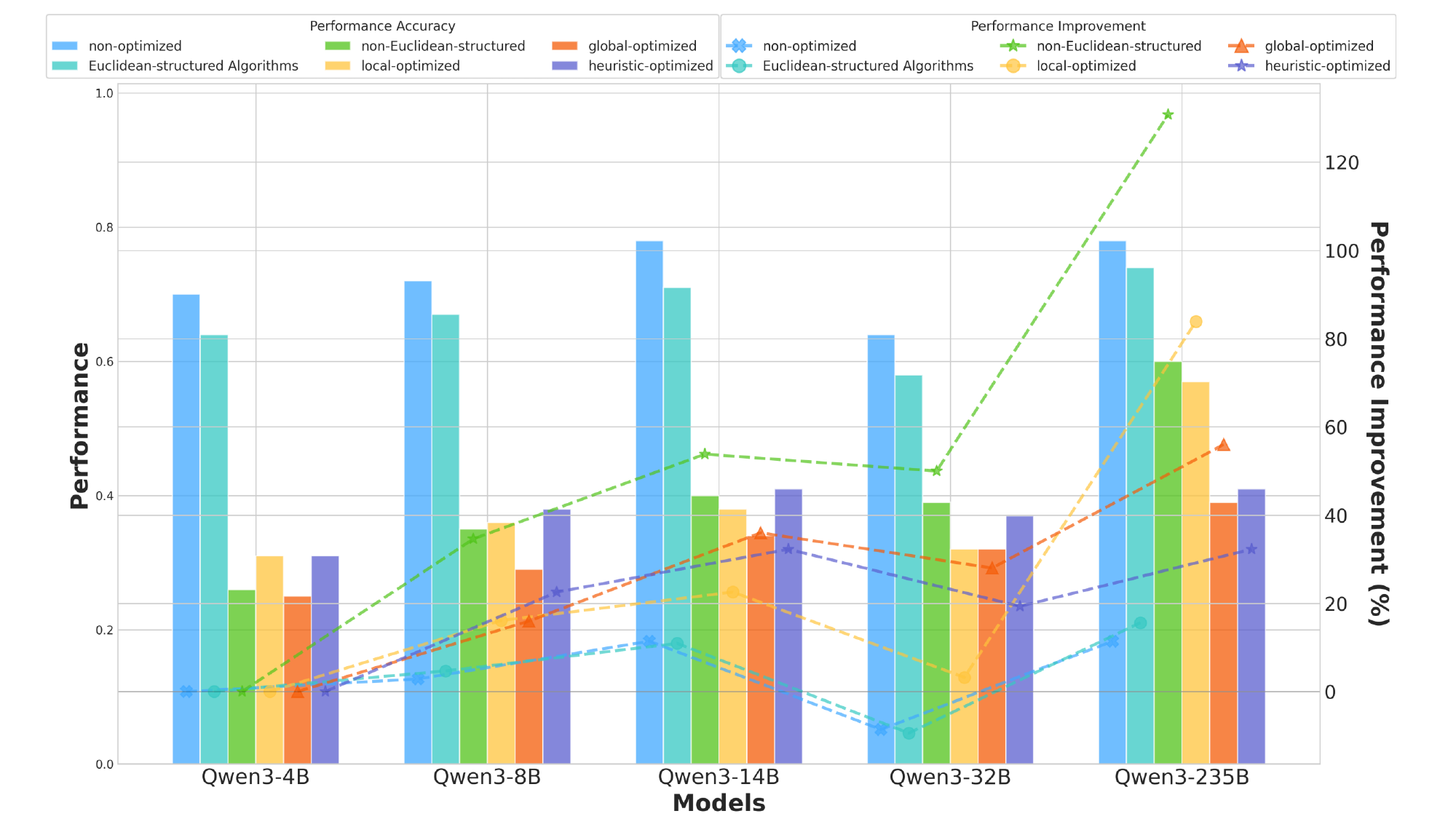}
  \caption{
    The results of models' parameters scaling.
}
  \label{fig: scaling_main}
\end{figure*}

\begin{figure*}[htbp]
    \centering
    \begin{subfigure}{0.48\textwidth}
        \centering
        \includegraphics[width=\linewidth]{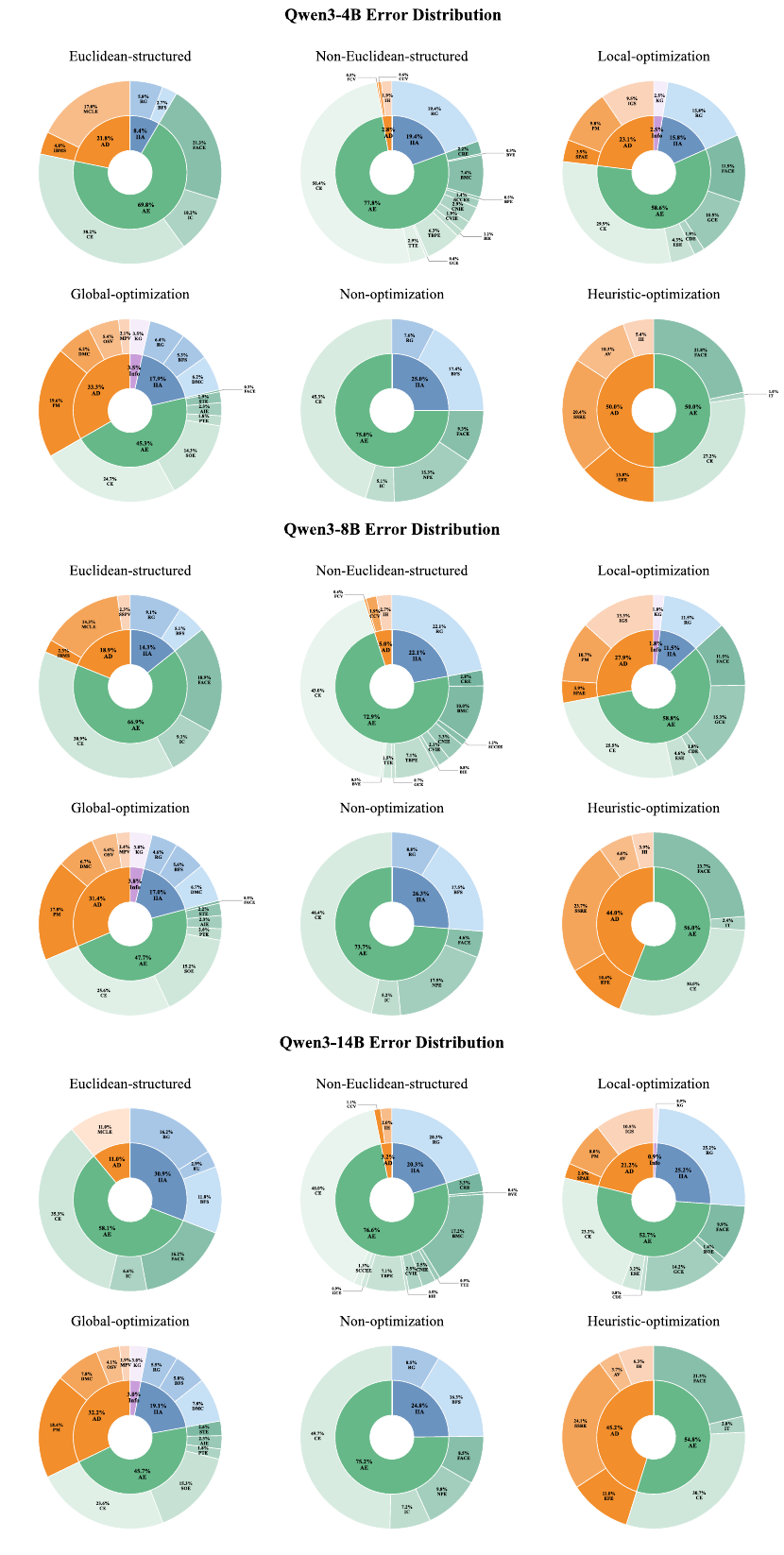}
        \label{fig:sub1}
    \end{subfigure}
    \hfill
    \begin{subfigure}{0.48\textwidth}
        \centering
        \includegraphics[width=\linewidth]{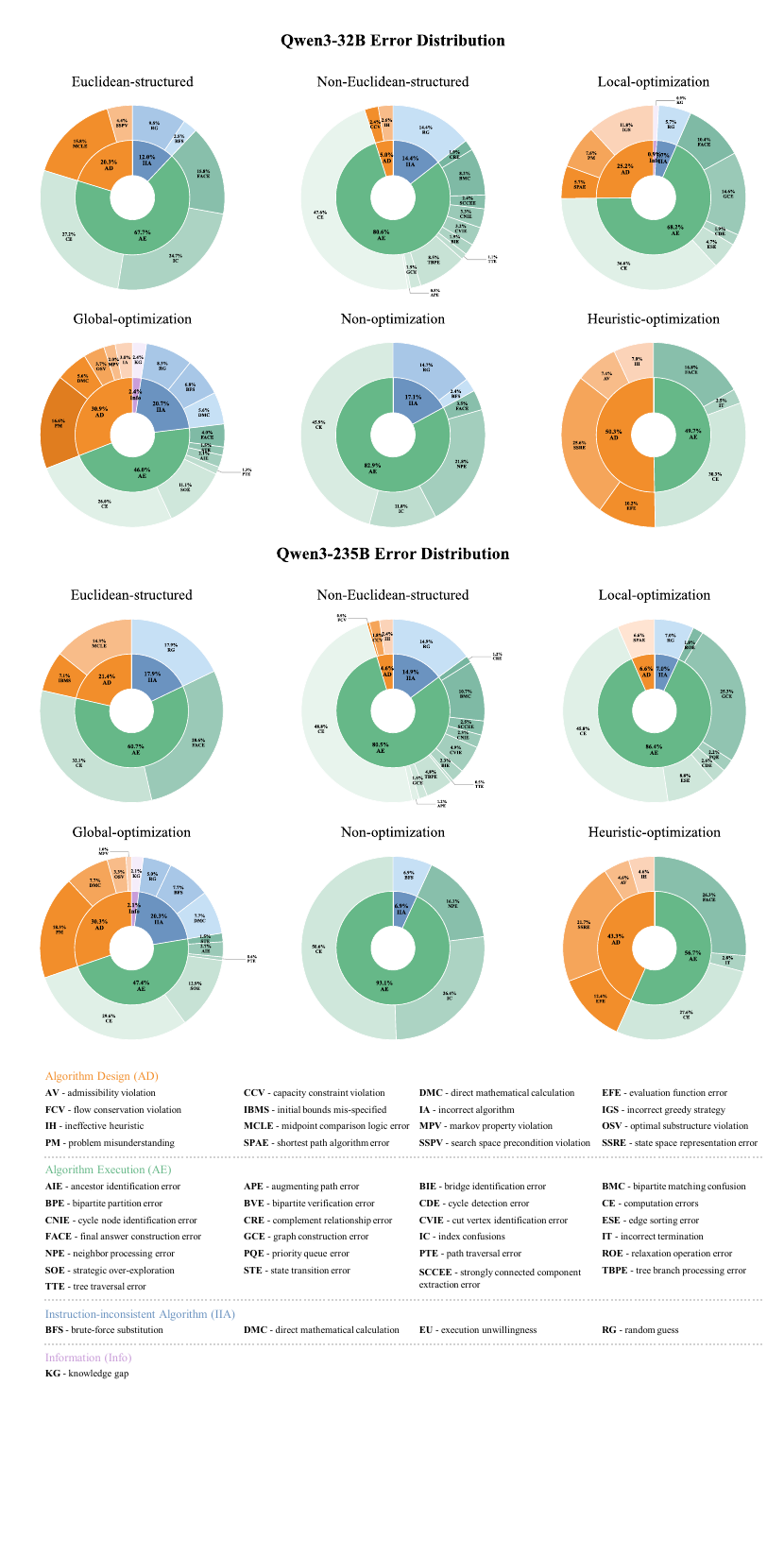}
        \label{fig:sub2}
    \end{subfigure}
    
    \caption{Error type distributions across algorithmic reasoning taxonomies from Qwen3-4B to Qwen3-235B.}
    \label{fig:total}
\end{figure*}

\begin{figure*}[htbp]
	\centering
  \includegraphics[width=\textwidth]{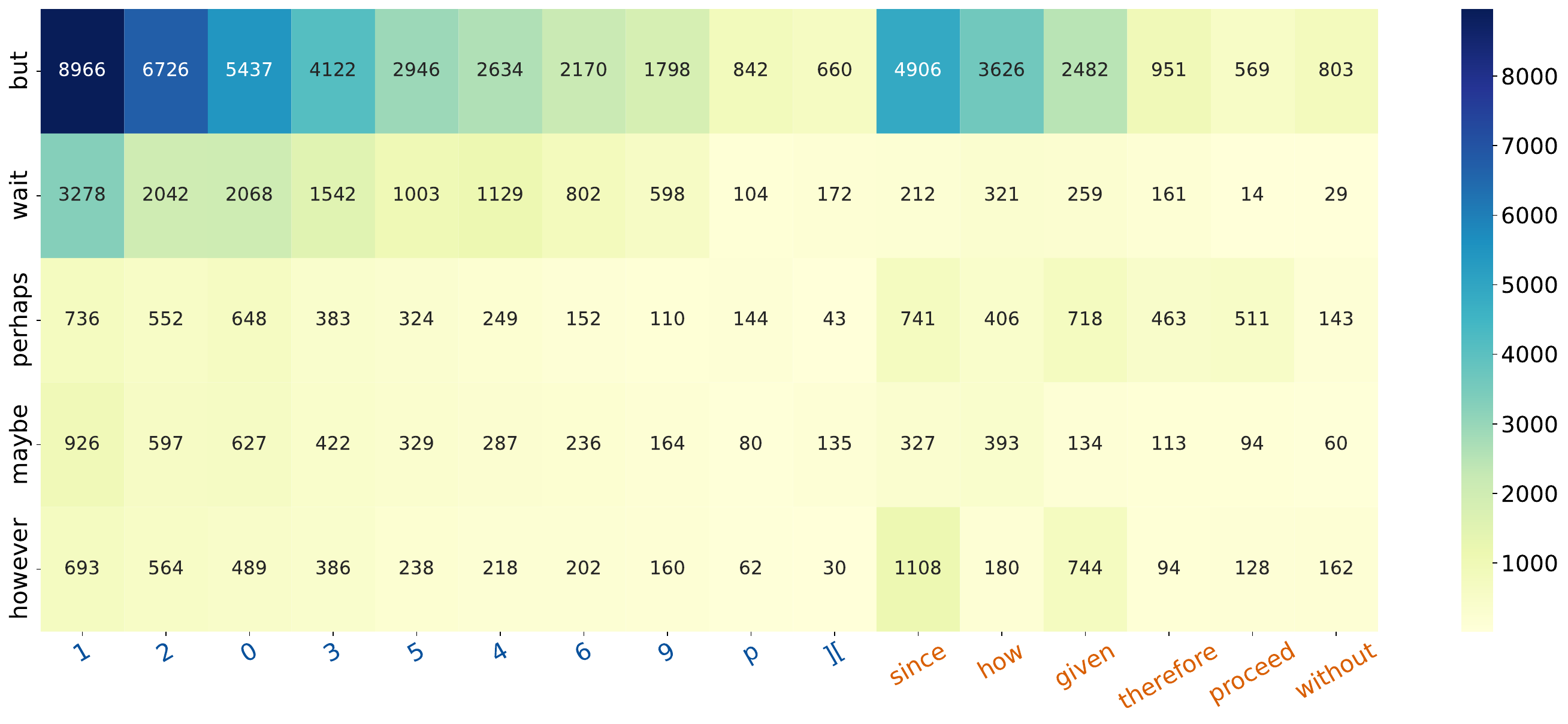}
  \caption{
    The co-occurrence heatmap of Qwen3-4B for low-entropy tokens proximal to indicator tokens (e.g., wait, but and so on) in global-optimized algorithmic reasoning. The blue segments along the x-axis denote necessary low-entropy tokens proximal to indicator tokens, whereas the orange segments correspond to regular high-entropy tokens appearing in the similar contextual positions. 
}
\label{fig: qwen3_4b_token_entropy}
\end{figure*}

\begin{figure*}[htbp]
	\centering
  \includegraphics[width=\textwidth]{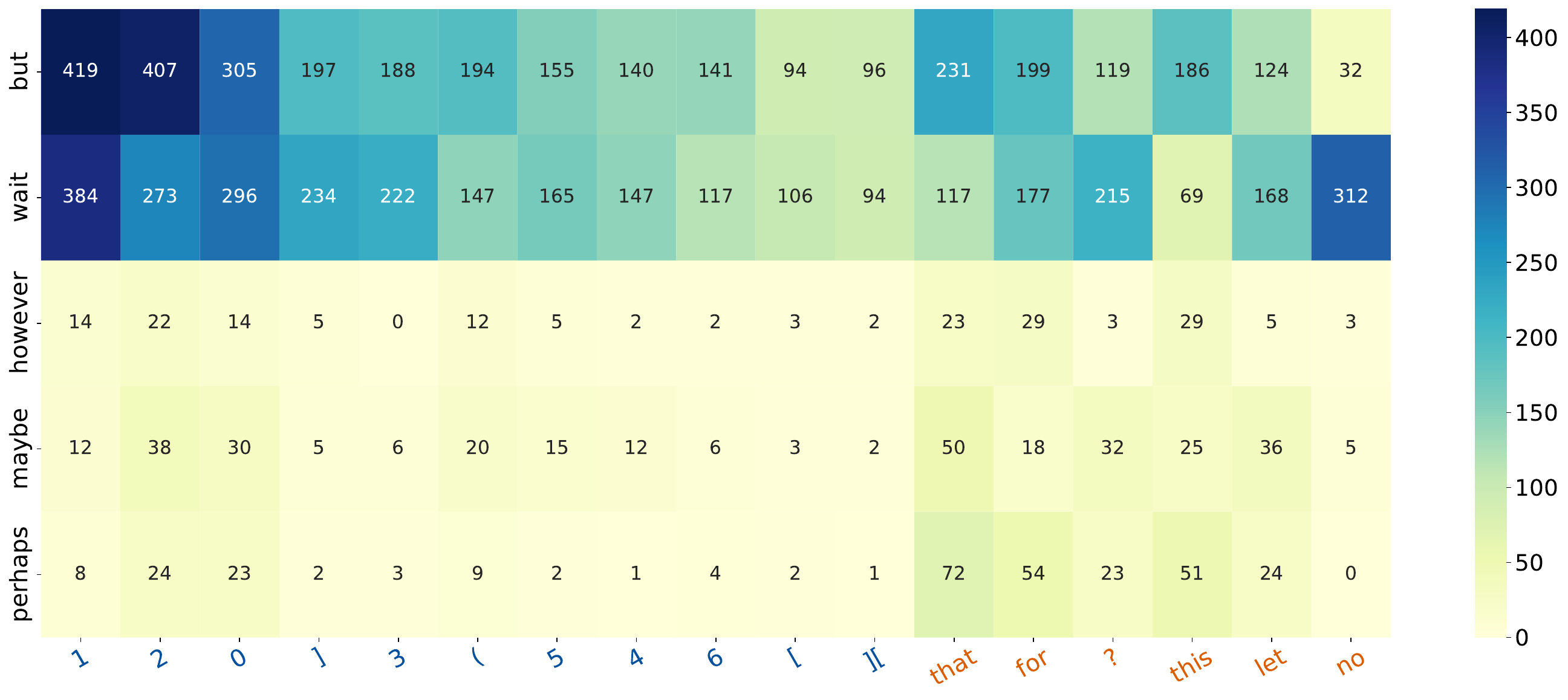}
  \caption{
    The co-occurrence heatmap of Qwen3-8B for low-entropy tokens proximal to indicator tokens (e.g., wait, but and so on) in global-optimized algorithmic reasoning. The blue segments along the x-axis denote necessary low-entropy tokens proximal to indicator tokens, whereas the orange segments correspond to regular high-entropy tokens appearing in the similar contextual positions.
}
\label{fig: qwen3_8B_token_entropy}
\end{figure*}

\begin{figure*}[htbp]
	\centering
  \includegraphics[width=\textwidth]{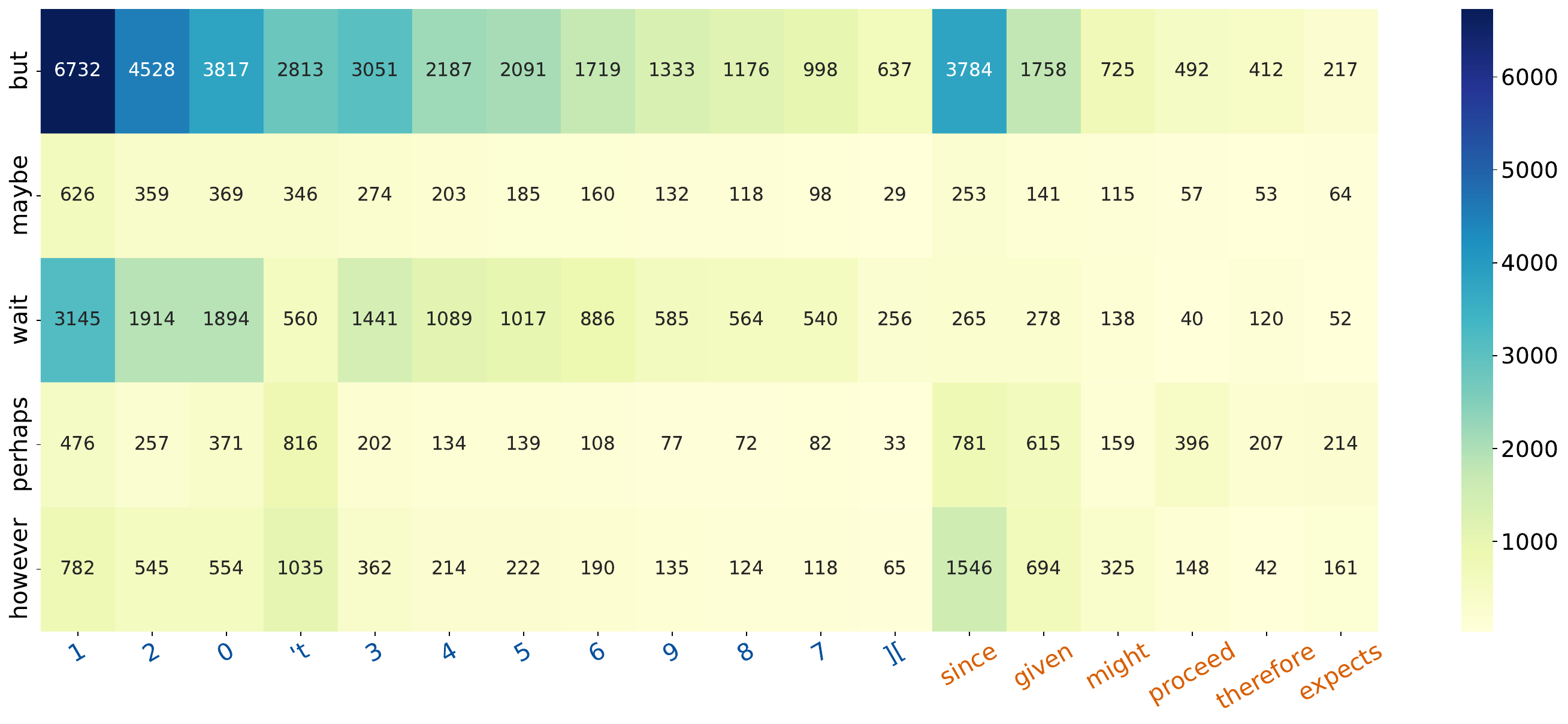}
  \caption{
    The co-occurrence heatmap of Qwen3-14B for low-entropy tokens proximal to indicator tokens (e.g., wait, but and so on) in global-optimized algorithmic reasoning. The blue segments along the x-axis denote necessary low-entropy tokens proximal to indicator tokens, whereas the orange segments correspond to regular high-entropy tokens appearing in the similar contextual positions.
}
\label{fig: qwen3_14B_token_entropy}
\end{figure*}

\begin{figure*}[htbp]
	\centering
  \includegraphics[width=\textwidth]{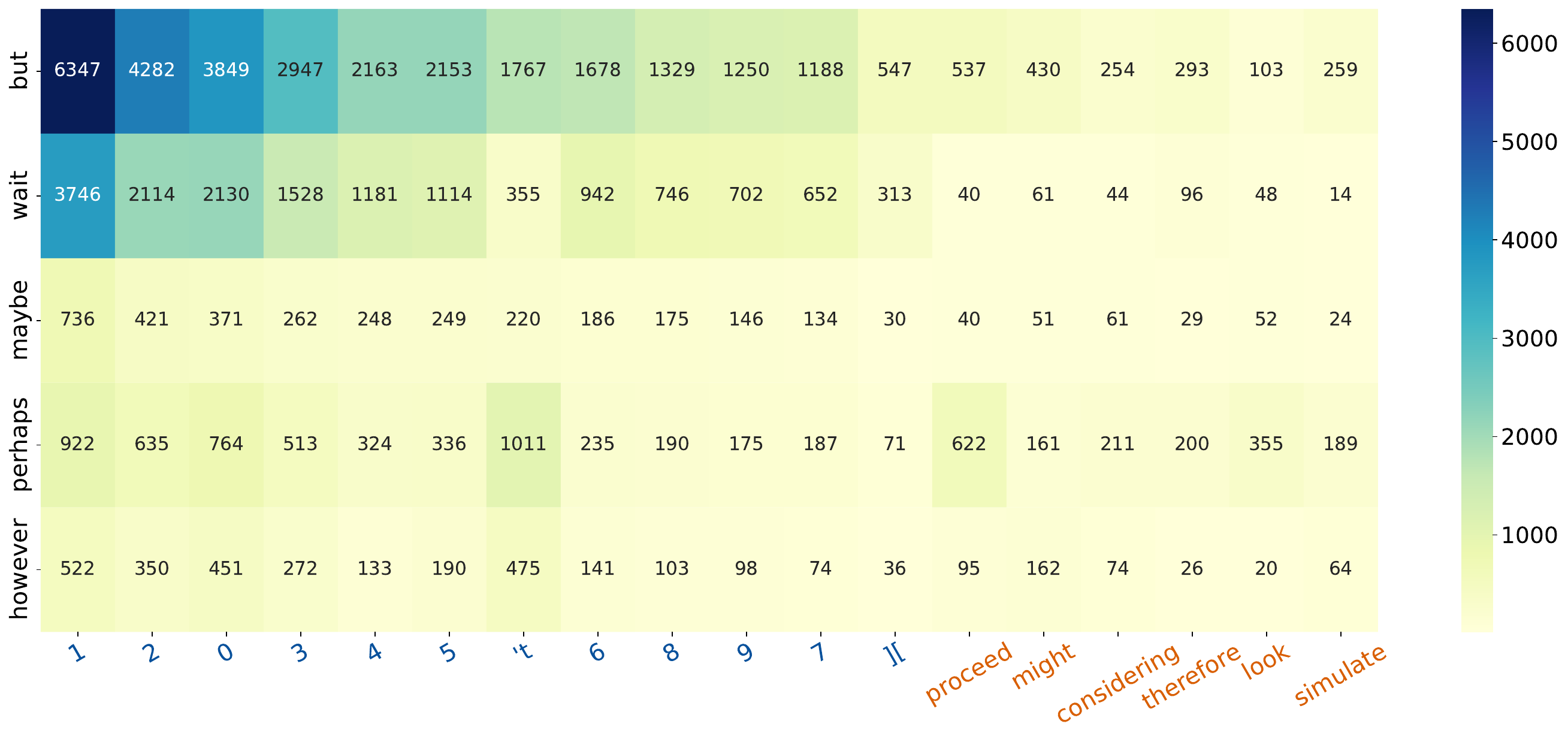}
  \caption{
    The co-occurrence heatmap of Qwen3-32B for low-entropy tokens proximal to indicator tokens (e.g., wait, but and so on) in global-optimized algorithmic reasoning. The blue segments along the x-axis denote necessary low-entropy tokens proximal to indicator tokens, whereas the orange segments correspond to regular high-entropy tokens appearing in the similar contextual positions.
}
\label{fig: qwen3_32B_token_entropy}
\end{figure*}

\begin{figure*}[htbp]
	\centering
  \includegraphics[width=\textwidth]{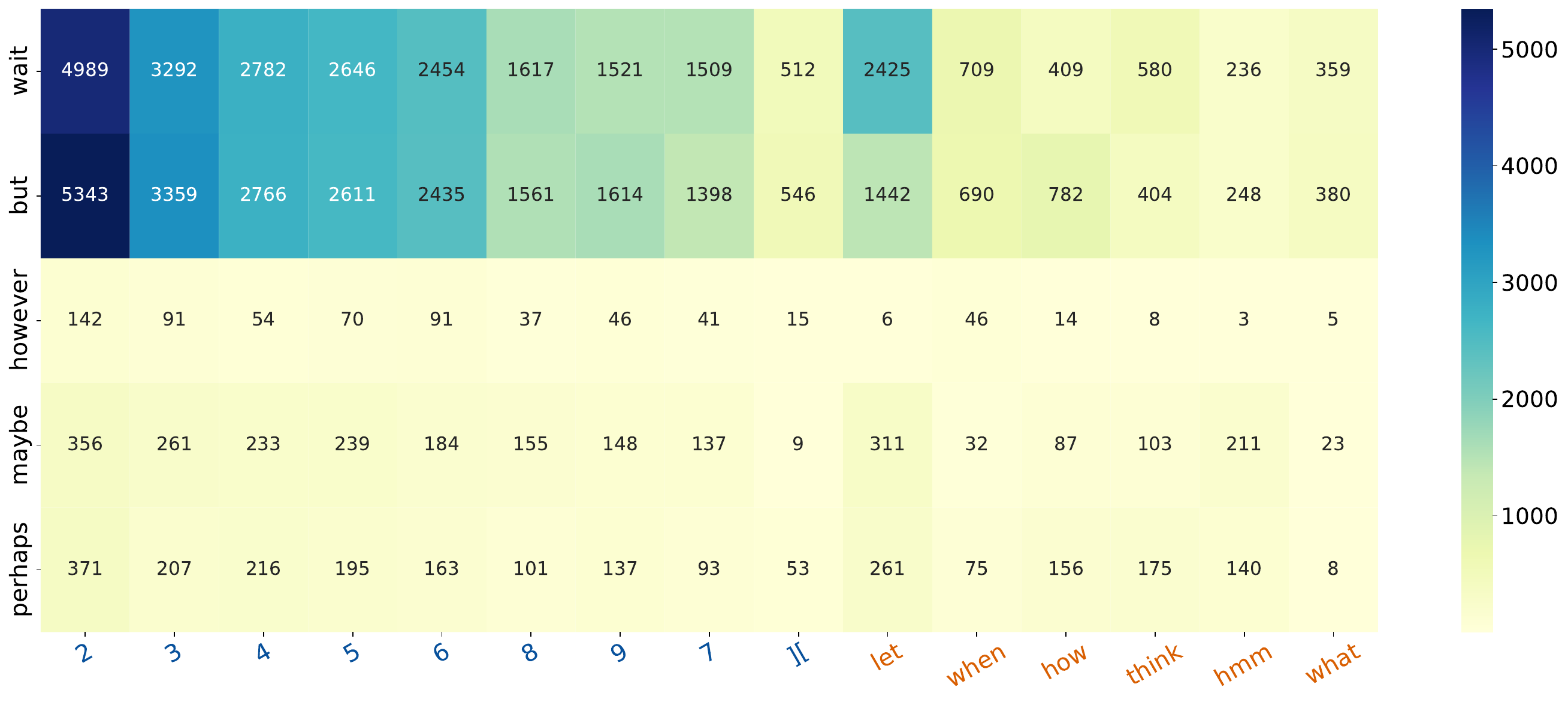}
  \caption{
    The co-occurrence heatmap of Qwen3-235B for low-entropy tokens proximal to indicator tokens (e.g., wait, but and so on) in global-optimized algorithmic reasoning. The blue segments along the x-axis denote necessary low-entropy tokens proximal to indicator tokens, whereas the orange segments correspond to regular high-entropy tokens appearing in the similar contextual positions.
}
\label{fig: qwen3_235B_token_entropy}
\end{figure*}

This section presents the additional experimental results in Section~\ref{sec: eval}, including Table~\ref{tab: structured_total}, Table~\ref{tab:non_heuristic_total}, Figure~\ref{fig: scaling_main}, Figure~\ref{fig:total}, Figure~\ref{fig: qwen3_4b_token_entropy}, Figure~\ref{fig: qwen3_8B_token_entropy}, Figure~\ref{fig: qwen3_14B_token_entropy}, Figure~\ref{fig: qwen3_32B_token_entropy} and Figure~\ref{fig: qwen3_235B_token_entropy}.

\begin{table*}[htbp]
\caption{Complete Models' performance on algorithms of structure-oriented, local-optimized and global-optimized ideas.}
\resizebox{\textwidth}{!}{
\begin{tabular}{@{}ccccc>{\columncolor{red!10}}c|ccccccc>{\columncolor{red!10}}c@{}}
\toprule
\multirow{2}{*}{LLMs} & \multicolumn{5}{c|}{Euclidean-structured algorithms}                         & \multicolumn{8}{c}{non-Euclidean-structured algorithms}                                                                                                                               \\ \cmidrule(l){2-14} 
                                                & ID & PS & Diff & BS & Avg.     & TDG & TUG & BGC & BGM & NF & DT & UG & Avg.         \\ \midrule
DeepSeek-v3.2-Speciale  &  0.77  &  0.93  &  0.93  & 0.89  &  0.88  &  0.73  &  0.72  &  0.58  &  0.59  &  0.58  &  0.90  &  0.83  &  0.70  \\
gemini-3-pro & 0.93  & 0.93  & 0.48  & 0.95  & 0.83  & 0.86  & 0.58  & 0.56  & 0.62  & 0.63  & 0.90  & 0.78  & 0.70  \\
grok4-vip                                       &        0.80             &      0.78      &     0.96       &        0.66       &    0.80     &           0.87                &           0.72                  &             0.76                 &        0.70                  &      0.62        &     0.96              &       0.80          &     0.78        \\
gpt-o3                                          &        0.63             &     0.95       &     0.48       &       0.94        &    0.75     & 0.87                     & 0.75                          & 0.44                    & 0.58                     & 0.62        & 0.85                & 0.70              & 0.69       \\
Qwen3-235B                                  & 0.63               & 0.95         & 0.50         & 0.89         & 0.74  & 0.81                     & 0.50                          & 0.42                    & 0.56                    & 0.50           & 0.76             & 0.68            & 0.60 \\
DeepSeek-R1                                     &       0.64              &     0.93       &     0.50       &       0.91        &     0.74    & 0.59                     & 0.49                       & 0.50                    & 0.53                    & 0.52        & 0.75             & 0.65           & 0.58 \\
gemini-2.5-pro                                  &          0.63           &    0.88        &      0.50      &       0.55        &    0.64     & 0.67                     & 0.53                        & 0.37                    & 0.58                     & 0.48         & 0.80                & 0.48           & 0.56 \\
Claude Opus                                     &      0.64               &      0.57     &     0.50       &      0.71         &   0.60      & 0.67                     & 0.57                       &               0.31           & 0.56                    &     0.40         &        0.73           &     0.35            &     0.51        \\
gpt-oss-120B                                    & 0.40                  & 0.60         & 0.42      & 0.63         & 0.51  & 0.54                     & 0.42                       & 0.26                    & 0.40                       & 0.27        & 0.58              & 0.35              & 0.40       \\ \midrule
Qwen3-4B                                        & 0.60                  & 0.93      & 0.40         & 0.62            & 0.64 & 0.24                     & 0.29                       & 0.23                     & 0.38                     & 0.26        & 0.32                  & 0.13            &      0.26       \\
Qwen3-8B                                        & 0.53               & 0.93      & 0.47      & 0.76            & 0.67 & 0.33                     & 0.43                       & 0.32                    & 0.46                    & 0.28        & 0.39             & 0.23            & 0.35       \\
Qwen3-14B                                       & 0.57               & 0.93      & 0.50         & 0.83         & 0.71 & 0.45                        & 0.43                        & 0.38                     & 0.48                    & 0.33         & 0.47             & 0.27           & 0.40 \\
Qwen3-32B                                       & 0.67               & 0.45         & 0.38      & 0.83         & 0.58 & 0.50                        & 0.48                       & 0.29                    & 0.48                    & 0.33        & 0.38             & 0.27           & 0.39 \\
QwQ-32B                                         & 0.57               & 0.95         & 0.37      & 0.83         & 0.68 & 0.62                     & 0.40                          & 0.33                     & 0.50                       & 0.41        & 0.62             & 0.34           & 0.46 \\
DeepSeek-R1-0528-Qwen3-8B                       & 0.50                  & 0.88      & 0.40         & 0.73         & 0.63   & 0.38                     & 0.45                          & 0.28                    & 0.43                     & 0.34        & 0.46             & 0.28          & 0.37 \\
Distill-Qwen-32B                                & 0.50                  & 0.95         & 0.45         & 0.76            & 0.67    & 0.50                        & 0.39                       & 0.19                    & 0.47                    & 0.38        & 0.45                & 0.22           & 0.37 \\
OpenThinker2-32B                                & 0.63               & 0.80         & 0.50         & 0.81         & 0.69    & 0.43                     & 0.41                       & 0.28                     & 0.43                     & 0.40           & 0.55                & 0.26           & 0.39 \\
Skyworker-OR1-32B                               & 0.67               & 0.93     & 0.50         & 0.86            & 0.74      & 0.56                     & 0.49                      & 0.33                   & 0.49                   & 0.40           & 0.58              & 0.29         & 0.45 \\
Qwen3-Coder-30B-A3B-Instruct                    & 0.47              & 0.80         & 0.22      & 0.57        & 0.51 & 0.12                    & 0.12                     & 0.08                    & 0.39                   & 0.17       & 0.21          & 0.07            & 0.17\\
Llama-3.3-70b-instruct                          & 0.17         & 0.62   & 0.12     & 0.38            & 0.32 & 0.07          & 0.08          & 0.03            & 0.17           & 0.03          & 0.19          & 0.10              & 0.09\\
gemma3-12B                                      & 0.33          & 0.40         & 0.0          & 0.39      & 0.28 & 0.10                        & 0.03                         & 0.09             & 0.30                       & 0.04       & 0.18      & 0.04     & 0.11\\
gemma3-27B                                      & 0.30                  & 0.50         & 0.08     & 0.49       & 0.34  & 0.11             & 0.08           & 0.06            & 0.29        & 0.08     & 0.22      & 0.08    & 0.13\\
Llama3-8B                                       & 0.07       & 0.07      & 0.0          & 0.13       & 0.07  & 0.05                         & 0.03           & 0.07        & 0.02               & 0.03      & 0.09              & 0.03           & 0.04 \\
Llama3.1-8B                                     & 0.07                & 0.13    & 0.0          & 0.20            & 0.10      & 0.03          & 0.03       & 0.06                     & 0.09                   & 0.02         & 0.08            & 0.03        & 0.05      \\
Mixtral-8x7B                                    & 0.07                & 0.03     & 0.0          & 0.15         & 0.06 & 0.08              & 0.01                        & 0.01                     & 0.09                     & 0.0            & 0.13              & 0.02            & 0.05 \\ 
\end{tabular}
}
\resizebox{\textwidth}{!}{
\begin{tabular}{@{}ccccc>{\columncolor{red!10}}c|ccccccc>{\columncolor{red!10}}c@{}}
\toprule
\multirow{2}{*}{LLMs}        & \multicolumn{5}{c|}{local-optimized algorithms} & \multicolumn{8}{c}{global-optimized algorithms} \\ \cmidrule(lr){2-6} \cmidrule(l){7-14}
                             & Greedy & SSSP  & MST   & DC    & Avg.  & LDP   & IDP   & TDP   & BLDP  & BDP   & MSSP  & LCA   & Avg.  \\ \midrule
DeepSeek-v3.2-Speciale       & 0.51   & 0.78  & 0.69  & 0.78  & 0.69  & 0.63  & 0.65  & 0.18  & 0.54   & 0.50  & 0.38  & 0.53  & 0.49  \\
gemini-3-pro                 & 0.54   & 0.73  & 0.61  & 0.63  & 0.63  & 0.73  & 0.30  & 0.18  & 0.52  & 0.27  & 0.39  & 0.61  & 0.43  \\
grok4-vip                    & 0.56   & 0.82  & 0.69  & 0.77  & 0.71  & 0.47  & 0.62  & 0.17  & 0.47  & 0.54  & 0.48  & 0.60  & 0.48  \\
gpt-o3                       & 0.52   & 0.78  & 0.64  & 0.78  & 0.68  & 0.59  & 0.47  & 0.20  & 0.56  & 0.47  & 0.40  & 0.48  & 0.45  \\
Qwen3-235B               & 0.57   & 0.75  & 0.53  & 0.43  & 0.57  & 0.59  & 0.34  & 0.22  & 0.52  & 0.38  & 0.37  & 0.33  & 0.39  \\
DeepSeek-R1                  & 0.47   & 0.69  & 0.47  & 0.63  & 0.56  & 0.54  & 0.43  & 0.14  & 0.32  & 0.39  & 0.40  & 0.40  & 0.38  \\
gemini-2.5-pro               & 0.42   & 0.68  & 0.46  & 0.59  & 0.54  & 0.41  & 0.44  & 0.16  & 0.53  & 0.33  & 0.34  & 0.39  & 0.37  \\
Claude Opus                  & 0.48   & 0.68  & 0.30  & 0.58  & 0.51  & 0.26  & 0.39  & 0.19  & 0.43  & 0.26  & 0.37  & 0.32  & 0.32  \\
gpt-oss-120B                 & 0.36   & 0.62  & 0.50  & 0.61  & 0.52  & 0.30  & 0.29  & 0.12  & 0.41  & 0.23  & 0.21  & 0.39  & 0.28  \\ \midrule
Qwen3-4B                     & 0.27   & 0.49  & 0.17  & 0.33  & 0.31  & 0.22  & 0.27  & 0.22  & 0.31  & 0.18  & 0.28  & 0.27  & 0.25  \\
Qwen3-8B                     & 0.32   & 0.51  & 0.26  & 0.38  & 0.36  & 0.37  & 0.27  & 0.30  & 0.32  & 0.22  & 0.28  & 0.31  & 0.29  \\
Qwen3-14B                    & 0.39   & 0.45  & 0.29  & 0.38  & 0.38  & 0.45  & 0.31  & 0.28  & 0.36  & 0.28  & 0.37  & 0.34  & 0.34  \\
Qwen3-32B                    & 0.34   & 0.52  & 0.22  & 0.18  & 0.32  & 0.35  & 0.35  & 0.27  & 0.33  & 0.17  & 0.38  & 0.40  & 0.32  \\
QwQ-32B                      & 0.33   & 0.58  & 0.19  & 0.53  & 0.41  & 0.49  & 0.31  & 0.19  & 0.40  & 0.28  & 0.33  & 0.37  & 0.34  \\
DeepSeek-R1-0528-Qwen3-8B    & 0.25   & 0.52  & 0.16  & 0.53  & 0.36  & 0.34  & 0.21  & 0.19  & 0.34  & 0.23  & 0.28  & 0.28  & 0.27  \\
Distill-Qwen-32B             & 0.27   & 0.43  & 0.24  & 0.42  & 0.34  & 0.37  & 0.27  & 0.09  & 0.31  & 0.20  & 0.29  & 0.31  & 0.26  \\
OpenThinker2-32B             & 0.31   & 0.57  & 0.25  & 0.40  & 0.38  & 0.33  & 0.22  & 0.11  & 0.35  & 0.23  & 0.32  & 0.36  & 0.27  \\
Skyworker-OR1-32B            & 0.33   & 0.50  & 0.38  & 0.48  & 0.42  & 0.44  & 0.29  & 0.08  & 0.36  & 0.24  & 0.31  & 0.33  & 0.29  \\
Qwen3-Coder-30B-A3B-Instruct & 0.15   & 0.31  & 0.04  & 0.28  & 0.20  & 0.06  & 0.12  & 0.18  & 0.13  & 0.14  & 0.15  & 0.10  & 0.13  \\
Llama-3.3-70b-instruct       & 0.19   & 0.13  & 0.01  & 0.08  & 0.10  & 0.06  & 0.09  & 0.09  & 0.18  & 0.15  & 0.08  & 0.08  & 0.10  \\
gemma3-12B                   & 0.12   & 0.16  & 0.0   & 0.17  & 0.11  & 0.08  & 0.05  & 0.11  & 0.09  & 0.14  & 0.12  & 0.06  & 0.10  \\
gemma3-27B                   & 0.17   & 0.18  & 0.03  & 0.22  & 0.15  & 0.17  & 0.13  & 0.08  & 0.14  & 0.17  & 0.21  & 0.03  & 0.13  \\
Llama3-8B                    & 0.05   & 0.13  & 0.0   & 0.13  & 0.08  & 0.0   & 0.05  & 0.0   & 0.02  & 0.0   & 0.02  & 0.0   & 0.01  \\
Llama3.1-8B                  & 0.02   & 0.08  & 0.0   & 0.11  & 0.05  & 0.03  & 0.03  & 0.06  & 0.03  & 0.03  & 0.03  & 0.01  & 0.03  \\
Mixtral-8x7B                 &     0.01   &   0.08    &   0.0    &   0.12    &  0.05     &    0.01   &  0.01     &   0.01    &   0.0    &   0.01    &    0.01   &    0.01   &   0.01    \\ \bottomrule
\end{tabular}}
\label{tab: structured_total}
\end{table*}

\begin{table*}[htbp]
\centering
\caption{Complete Models' performance on algorithms of structure-oriented ideas.}
\resizebox{0.9\textwidth}{!}{
\begin{tabular}{@{}cccc>{\columncolor{red!10}}c|cc>{\columncolor{red!10}}c@{}}
\toprule
\multirow{2}{*}{LLMs}        & \multicolumn{4}{c|}{non-optimized algorithms} & \multicolumn{3}{c}{heuristic-optimized algorithms} \\ \cmidrule(lr){2-5} \cmidrule(l){6-8}
                             & BO    & Recursion & Searching & Avg.  & AS    & IDAS    & Avg.  \\ \midrule
DeepSeek-v3.2-Speciale       & 0.97  & 0.93      & 0.87      & 0.92  & 0.39  & 0.58  & 0.49  \\
gemini-3-pro                 & 0.97  & 0.91      & 0.78      & 0.89  & 0.34  & 0.25  & 0.30  \\
grok4-vip                    & 0.90  & 0.53      & 0.87      & 0.77  & 0.40  & 0.39  & 0.39  \\
gpt-o3                       & 1.0   & 0.89      & 0.80      & 0.90  & 0.32  & 0.47  & 0.39  \\
Qwen3-235B               & 0.80  & 0.78      & 0.77      & 0.78  & 0.38  & 0.45  & 0.41  \\
DeepSeek-R1                  & 0.85  & 0.86      & 0.75      & 0.82  & 0.41  & 0.65  & 0.53  \\
gemini-2.5-pro               & 0.84  & 0.87      & 0.69      & 0.80  & 0.24  & 0.26  & 0.25  \\
Claude Opus                  & 0.85  & 0.89      & 0.61      & 0.79  & 0.37  & 0.55  & 0.46  \\
gpt-oss-120B                 & 0.47  & 0.37      & 0.60      & 0.48  & 0.26  & 0.30  & 0.28  \\ \midrule
Qwen3-4B                     & 0.82  & 0.86      & 0.41      & 0.70  & 0.22  & 0.40  & 0.31  \\
Qwen3-8B                     & 0.93  & 0.91      & 0.33      & 0.72  & 0.28  & 0.48  & 0.38  \\
Qwen3-14B                    & 0.87  & 0.94      & 0.53      & 0.78  & 0.28  & 0.54  & 0.41  \\
Qwen3-32B                    & 0.73  & 0.70      & 0.50      & 0.64  & 0.24  & 0.50  & 0.37  \\
QwQ-32B                      & 0.80  & 0.72      & 0.33      & 0.62  & 0.27  & 0.55  & 0.41  \\
DeepSeek-R1-0528-Qwen3-8B    & 0.73  & 0.77      & 0.54      & 0.68  & 0.27  & 0.53  & 0.40  \\
Distill-Qwen-32B             & 0.81  & 0.83      & 0.48      & 0.71  & 0.31  & 0.44  & 0.38  \\
OpenThinker2-32B             & 0.82  & 0.81      & 0.51      & 0.71  & 0.28  & 0.52  & 0.40  \\
Skyworker-OR1-32B            & 0.73  & 0.72      & 0.55      & 0.67  & 0.33  & 0.51  & 0.42  \\
Qwen3-Coder-30B-A3B-Instruct & 0.63  & 0.31      & 0.19      & 0.38  & 0.16  & 0.29  & 0.23  \\
Llama-3.3-70b-instruct       & 0.47  & 0.69      & 0.20      & 0.45  & 0.10  & 0.32  & 0.21  \\
gemma3-12B                   & 0.37  & 0.63      & 0.16      & 0.38  & 0.08  & 0.31  & 0.19  \\
gemma3-27B                   & 0.50  & 0.77      & 0.23      & 0.50  & 0.16  & 0.29  & 0.23  \\
Llama3-8B                    & 0.17  & 0.06      & 0.10      & 0.11  & 0.02  & 0.06  & 0.04  \\
Llama3.1-8B                  & 0.13  & 0.33      & 0.08      & 0.18  & 0.06  & 0.13  & 0.10  \\
Mixtral-8x7B                 & 0.17  & 0.11      &    0.03       &  0.07     &   0.07    &  0.13     &   0.10    \\ \bottomrule
\end{tabular}}
\label{tab:non_heuristic_total}
\end{table*}

\end{document}